\documentclass[letterpaper, 10 pt, journal, twoside]{IEEEtran}

\usepackage{epsfig}
\usepackage{amssymb}
\usepackage{cite}
\usepackage{booktabs}
\usepackage{multirow}
\usepackage{mathtools}
\usepackage{bm}

\newcommand{\figlab}[1]{\label{fig:#1}}
\newcommand{\figref}[1]{Fig.~\ref{fig:#1}} 
\newcommand{\tablab}[1]{\label{tab:#1}}
\newcommand{\tabref}[1]{Table~\ref{tab:#1}} 
\newcommand{\forlab}[1]{\label{for:#1}}

\newcommand{\algolab}[1]{\label{algorithm:#1}}
\newcommand{\algoref}[1]{Algorithm~\ref{algorithm:#1}} 

\newcommand{\ie}{\textit{i.e.,}}
\newcommand{\eg}{\textit{e.g.,}}
\newcommand{\etal}{\textit{et~al.}}

\usepackage{subcaption}
\usepackage{url}
\usepackage{threeparttable}

\usepackage{algorithmicx}
\usepackage[ruled]{algorithm}
\usepackage[noend]{algpseudocode}

\begin{document}

\title{Robotic Waste Sorter with Agile Manipulation and Quickly Trainable Detector}

\author{Takuya Kiyokawa,~\IEEEmembership{Member,~IEEE}, Hiroki~Katayama, Yuya~Tatsuta\\
Jun~Takamatsu,~\IEEEmembership{Member,~IEEE}, and Tsukasa~Ogasawara,~\IEEEmembership{Member,~IEEE}
\thanks{All authors are with Division of Information Science, Nara Institute of Science and Technology, 8916-5 Takayama-cho, Ikoma-shi, Nara, Japan.}
\thanks{$^{1}$Corresponding author's email: kiyokawa.takuya@is.naist.jp}}

\markboth{Journal of \LaTeX\ Class Files,~Vol.~X, No.~X, XXXX~2020}
{Kiyokawa \MakeLowercase{\textit{et al.}}: Robotic Waste Sorter with Agile Manipulation and Quickly Trainable Detector}

\maketitle

\begin{abstract}
Owing to human labor shortages, the automation of labor-intensive manual waste-sorting is needed. The goal of automating waste-sorting is to replace the human role of robust detection and agile manipulation of waste items with robots. 
To achieve this, we propose three methods. 
First, we provide a combined manipulation method using graspless push-and-drop and pick-and-release manipulation.
Second, we provide a robotic system that can automatically collect object images to quickly train a deep neural--network model. 
Third, we provide a method to mitigate the differences in the appearance of target objects from two scenes: one for dataset collection and the other for waste sorting in a recycling factory. 
If differences exist, the performance of a trained waste detector may decrease.
We address differences in illumination and background by applying object scaling, histogram matching with histogram equalization, and background synthesis to the source target-object images. 
Via experiments in an indoor experimental workplace for waste-sorting, we confirm that the proposed methods enable quick collection of the training image sets for three classes of waste items (\ie~aluminum can, glass bottle, and plastic bottle) and detection with higher performance than the methods that do not consider the differences. We also confirm that the proposed method enables the robot quickly manipulate the objects. 
\end{abstract}

\begin{IEEEkeywords}
Robotics and automation, robot vision systems, computer vision, recycling, machine learning, object detection.
\end{IEEEkeywords}

\section{Introduction}
\label{sec:introduction}
In the context of long-standing human-labor shortages, the automation of various tasks by robots is ever more in demand. 
The automation of sorting container and packaging waste is an urgent example, and several related studies have been conducted worldwide~\cite{Gundupalli2017,Chahine2018,Industry4Colombia,ReviewWasteRobot}. 
Among the general waste articles produced by society, container and packaging wastes are dominant. 
Thus, many companies have been tackling this issue~\cite{RecyclingRobots,Lukka2014}.

Normally, vast amounts of unsorted recyclable waste are gathered at a collection site and manually sorted into designated boxes or transport lanes according to categories (\eg~aluminum can, glass bottle, or plastic bottle). 
The goal of automating this process is to replace the human role of detection and manipulation of the waste items with robots. 

A key difficulty is agility, because conveyor transportation speeds should be as high as possible, owing to the large volumes of waste to be sorted.
Another challenge is to robustly detect short lifecycle objects that are dirty on the surface or deformed and/or damaged.

With this in mind, we construct a robotic waste-sorting system (see~\figref{system}) with the robust detection and agile manipulation needed for recycling factories.
In this study, to achieve agile waste-sorting manipulation, we first propose a combined manipulation method using graspless \textit{push-and-drop} and \textit{pick-and-release} manipulation. 
Second, we propose a robotic training dataset collection system to automatically capture images and annotate them for training a deep--learning (DL)-based waste detector. 
We attempt to improve the robustness by applying a domain adaptation method to the collected dataset.

DL-based object detectors~\cite{FastRCNN,FasterRCNN,SSD,YOLOv3,M2Det,FCOS,EfficientDet} can infer the location and category of objects having a variety of appearances in images. 
However, massive training datasets~\cite{ImageNet,PascalVOC, MSCOCO,CityScapes} are required, owing to the many parameters to be optimized~\cite{ReviewDL}. 
With recent decreases in product lifecycles, unknown waste items frequently appear at the sorting factories. 
Thus, we must quickly update the training dataset with new waste images for fine-tuning.

To quickly create an object-image dataset using our system, a target object is placed on an automatic rotating stage and imaged from multiple viewpoints using a hand--eye robot arm shown, as in~\figref{collection-system}. 
The robot arm and rotating stage are automatically controlled while capturing images.
Our previously proposed automatic annotation method~\cite{RA-L2019} using augmented-reality (AR) marker detection~\cite{Kato2001} is applied to captured images. 
To train the DL-based waste detector, we place the collection-target object on the rotating table for image capture. However, we do not have to manually annotate the images.
Using automatic annotation methods of this nature, prior experiments have achieved a six-class object detection~\cite{AdvancedRobotics2019}.

\begin{figure}[tb]
  \centering
  \includegraphics[width=0.98\linewidth]{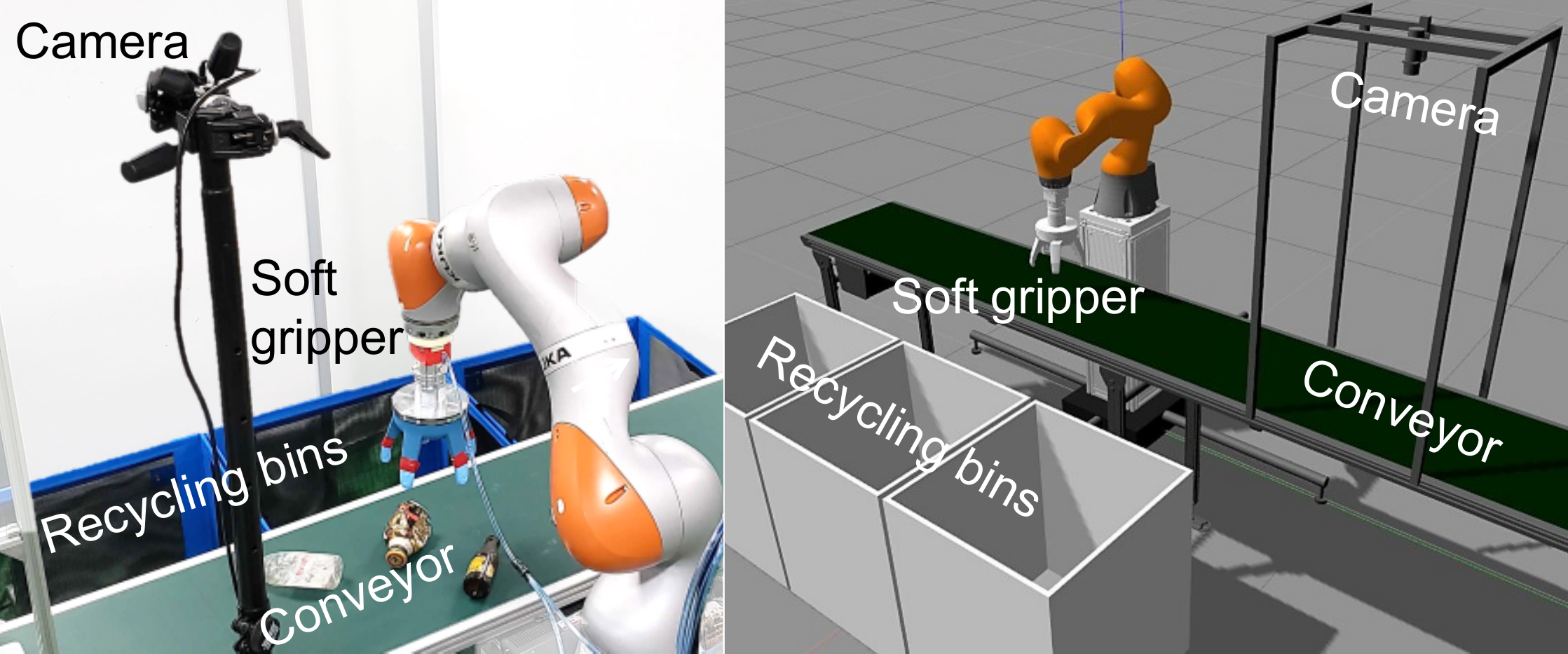}
  \caption{\small Configuration of proposed automated waste-sorting system.}
  \figlab{system}
\end{figure}
\begin{figure}[tb]
  \centering
  \includegraphics[width=0.9\linewidth]{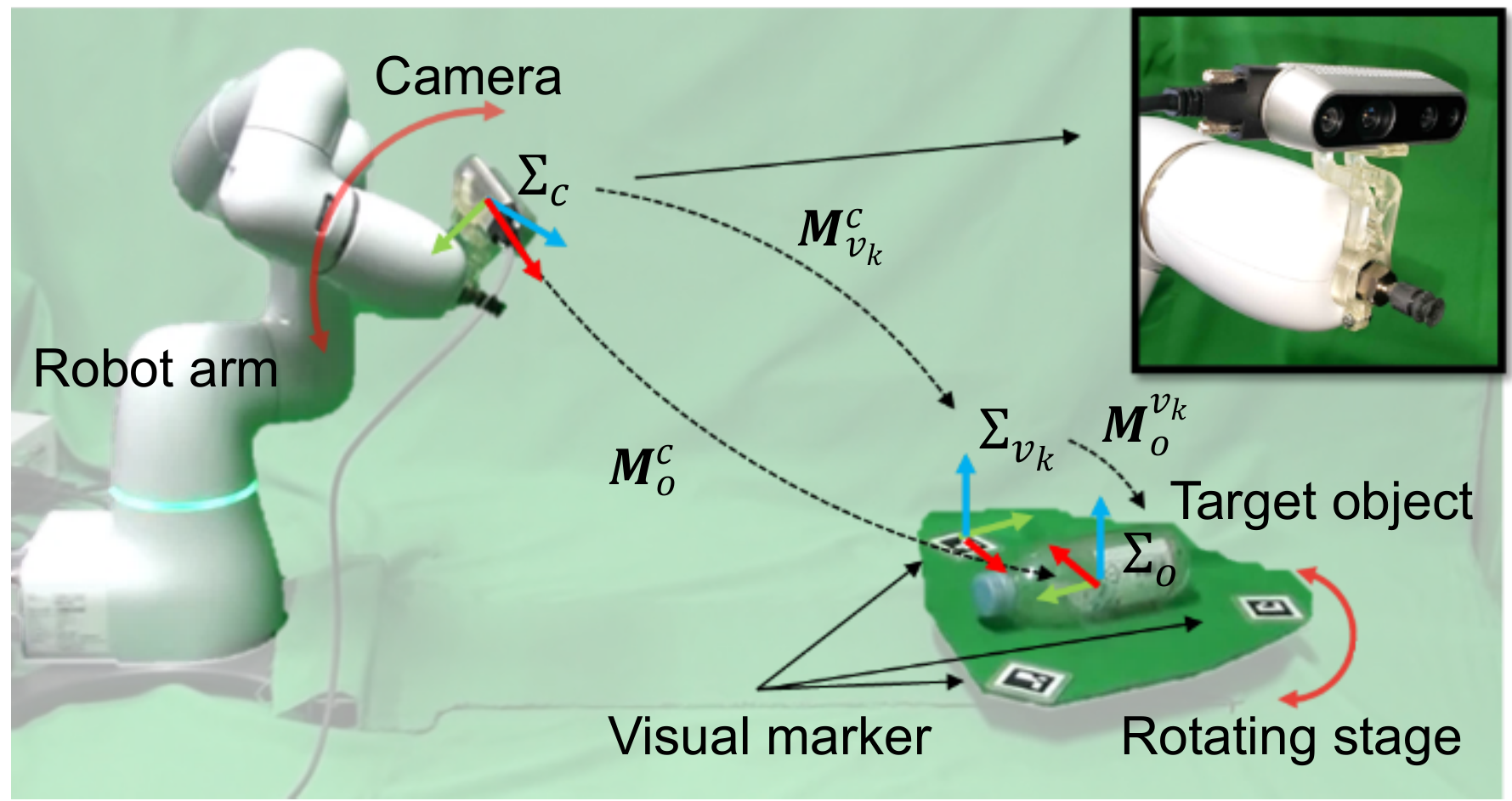}
  \caption{\small Robotic training dataset collection system that facilitates image capturing and automatically annotates labels and bounding boxes.}
  \figlab{collection-system}
\end{figure}
Although object images in the real-world can be easily provided, they often appear differently from items found in the working environment. 
Thus, detection performance can decrease when collecting images without consideration for adaptation methods. 

The waste-sorting workplace exists in an indoor environment for this study. 
Thus, it can be fixed in terms of illumination and background. 
We propose methods to reduce the differences easily and effectively for such conditions.

This study focuses on two domain differences in terms of illumination and background between the dataset collection environment in~\figref{collection-system} and the waste-sorting environment in~\figref{system}. 
First, we adjust the object size in the image to be as close as possible to the real one in the waste-sorting scene.
Subsequently, we apply histogram matching (HM) to images using a red--green--blue (RGB) color space to reduce illumination differences. 
Based on our qualitative observations for RGB histograms of the object images captured in the waste-sorting environment, we apply histogram smoothing for the collected images to further make the RGB histogram resemble the destination images.
Furthermore, to reduce the differences of background conditions, we use background-synthesized and histogram-matched images as the training images.

The contributions of this study are threefold. 
\begin{enumerate}
  \item In the proposed sorting manipulation method by push-and-drop, the time required for the manipulation of one object is about 1.9~s faster than pick-and-release.
  \item The proposed robotic training dataset collection system composed of a hand--eye robot arm, a rotating stage, and visual markers enables agile object-image capturing from multiple viewpoints. The time required for the proposed automatic collection is 12.3~s: 99.1\% faster than prior methods.
  \item As a benefit of proposed object-image dataset adaptation method, we achieve improved waste-detection accuracy. We further propose the addition of a small real-world dataset captured in the waste-sorting scene to the domain-adapted dataset. Training with this dataset achieves a detection accuracy of 79\%, which is 39\% higher than using the original one that lacks domain adaptation and real-world images.
\end{enumerate}

\section{Related Work}
\subsection{Robotic Waste Sorter}
To achieve an agile robotic sorter for a huge volume of waste, previous studies sorted items transported on a conveyor using suction grippers for quick grasping and manipulation~\cite{Lukka2014, ZenRobot2015}. 
Graspless~\cite{GrasplessAiyama,GrasplessMaeda}, prehensile pushing~\cite{Nikhil2015}, and non-prehensile manipulation~\cite{NonprehensileManip,DynamicNonPrehensile} methods, like our push-and-drop technique, have not been applied thus far. Therefore, the feasibility of push-and-drop has remained untested until now, notwithstanding that such manipulations using robotic hands are reasonable methods of agile manipulation.

Conventional automatic sorting systems are based on different types of sensors (\eg~optical~\cite{Plastics,OpticalSensor,LaserSensor} and thermal techniques~\cite{Thermalewaste,ThermalMetal}). Mao~\etal~\cite{Mao2021} proposed a classifier using a convolutional neural network to classify an RGB object image that included one waste item.
Furthermore, DL-based algorithms using RGB and RGB-depth (RGBD) sensors have been used to detect and segment individual waste items from a densely cluttered pile~\cite{Lukka2014,ZenRobot2015,BinYan2017,ButtonCellBatteries,Zhifei2019}. 

\subsection{Generating a Training Dataset for a DL-based Detector}
Deep convolutional neural networks can automatically discover the needed representations for object detection and classification from large datasets in a manner similar to that of the human visual cortex~\cite{DeepLearning}. 
Although larger datasets enable robust detection and classification of waste items having diverse appearances, the construction of such datasets demands an enormous amount of time and effort. 
Binyan \etal~\cite{BinYan2017} used 47,988 images of recyclable waste on a conveyor for training and testing a deep neural--network model. 
3,999 images were originally collected, and additional ones were augmented via flipping and scaling the collected images. 
Bai \etal~\cite{OnGlass} achieved garbage recognition with small errors using training datasets comprising 40,000 training and 7,000 testing images grouped into six classes: five garbage and one non-garbage. 
Zhihong \etal~\cite{GarbageDetChina} used 1,480 images only for the detection of a glass bottle on a conveyor transporting various waste items.
These are distinguished from automatic collection methods like ours.

DL-based vision systems are fast and can detect vast categories of objects. However, as mentioned, the cost of manual image annotation remains very high.
To tackle this, two major efforts to easily collect large datasets are under way. 
One approach includes (1) data augmentation to enrich image datasets for improving the generalizability of DL models, and the other deals with (2) the simplification of labor-intensive annotation processes to increase the number of datasets with reduced human intervention. This study applies both types.

In the research of (1), Takahashi \etal~\cite{RICAP} applied random-image cropping and patching to improve classification accuracy. 
Zhong \etal~\cite{RondomErasing} applied random erasing to reduce the risk of over-fitting and made the model robust to occlusion. 
They randomly changed pixel intensities within the selected region of an arbitrary size. 
Cubuk \etal~\cite{AutoAugment} proposed a method of automatically searching for data augmentation policies directly from a dataset (\textit{AutoAugment}). 
Each policy expresses several choices and orders of possible augmentation operations, wherein each operation is an image--processing function (\eg~translation, rotation, or color normalization). 
Lim \etal~\cite{FastAutoAugment} proposed \textit{FastAutoAugment}, an improved policy extraction method that is significantly faster than the original AutoAugment, which requires thousands of graphical-processing-unit hours, even for small datasets.

In the research of (2), to make human annotation easier, effective and easy-to-use annotation tools~\cite{ExClick,DeepExCut} were proposed. 
However, with these, humans still spent too much time on annotation. 
For example, polygonal annotations for instance segmentation were conducted via interactive image-region mouse clicks by human annotators. 
Huan \etal~\cite{NVIDIAannotation} proposed a graph-convolutional network, \textit{Curve-GCN}, to automatically predict the vertices of instances in the images. 
An annotator can choose any wrong control points and move them onto the correct object boundary. 
Only its immediate neighbors will be re-predicted based on manual annotation.
Rodrigo \etal ~\cite{GoogleAnnotation} designed software capable of correcting wrong annotations by clicking on images. 
Based on the corrective clicks, the segmentation mask for the annotation was automatically updated.
These human-in-the-loop polygonal annotations take only a few seconds for each image, but they also require corrective clicks for the vertices, owing to the need for annotation quality assurance. 

Another interesting approach is the use of an RGBD sensor~\cite{EasyLabel} and visual markers~\cite{SemiautoLabel,SemiautoTrainGen} to automatically segment objects from the background. 
These approaches are like ours. However, in the previous approaches, the automatic collection of multi-view object images and their domain adaptations were out-of-scope. 
Our robotic training dataset collection system of multi-view images gives the dataset variety and quantity and is useful when training the garbage detector to handle various appearances.
Image adaptation methods of reducing the differences of domains are necessary to enable faster image collection.

\subsection{Domain Adaptation for DL-based Vision System}
Despite the many ideas explored, the predominant datasets were built by humans using bounding boxes or polygonal masks~\cite{ImageNet,PascalVOC, MSCOCO,CityScapes}.
Our proposed method can automatically annotate object images without human intervention. 
Because there are differences in object appearance between the dataset collection environment shown in~\figref{collection-system} and the waste-sorting environment shown in~\figref{system}, the collected dataset using the robotic collection system could not be directly used to train the waste detector.

Domain adaptation is a specific scenario in transfer learning that can be used to effectively remove domain differences. 
Domain adaptation has been shown to be effective for the transfer learning of models in different computer vision tasks, including image classification~\cite{Tzeng2017}, object recognition~\cite{Raghuraman2011}, object detection for indoor kitchen scenes~\cite{SynthesizingRSS17}, outdoor scenes~\cite{Hsu2019}, water-colors~\cite{Inoue2018}, and semantic segmentation~\cite{Luo2019}. 

Georgakis \etal~\cite{SynthesizingRSS17} tackled an issue like ours. 
To automatically generate image datasets that emulate real environments, they superimposed two-dimensional images of textured object models into images of real indoor environments reflecting a variety of locations and scales.
They verified the efficacy of a seamless cloning (SC) method to mitigate the effects of changes in illumination and contrast. 
They also verified an object--scaling method that used the depth of the selected position of a real household environment.

In this study, we tackle the issue of domain adaptation for a collected waste-image dataset ourselves so that it can be adapted to a real waste-sorting problem. 
For this reason, we create a waste dataset using images of 33 aluminum cans, 33 glass bottles, and 33 plastic bottles. 

We also strongly support the efficacy of domain adaptation for the waste-sorting environment.
In particular, we evaluate more methods to mitigate the changes of object-size appearance, image illumination, contrast and background.

\section{Automatically Generating Training Dataset}
This section first describes the proposed robotic training dataset collection system using a small hand--eye robot arm and an automatic rotating stage. 
Next, we explain the methods for reducing the differences of the illumination and the background. 
The object appearances differ between dataset-collection and waste-sorting environments.

For domain adaptation, we consider how to match the original domain of the generated training dataset to that of the target domain of the waste-sorting environment.

\subsection{Multi-viewpoint Object Image Acquisition}
\figref{collection-system} shows our robotic training dataset collection system that includes a small hand--eye robot arm and a controllable rotating stage. 
Using the small hand--eye robot arm equipped with an RGB camera, we collect images from multiple viewpoints by moving the robot arm to capture a target object placed on the automatic rotating stage. 

An RGBD camera is used for both object-image dataset collection and the robot vision capability of the proposed robotic waste-sorting system, because we minimize the effects of the camera in the detection experiments. 
Depth information is not used to generate the training dataset, but the same camera as the waste-sorting environment is.
The white balance and the exposure of the camera are fixed during image dataset collection and robot experiments.
\begin{figure}[tb]
  \centering
  \includegraphics[width=0.55\linewidth]{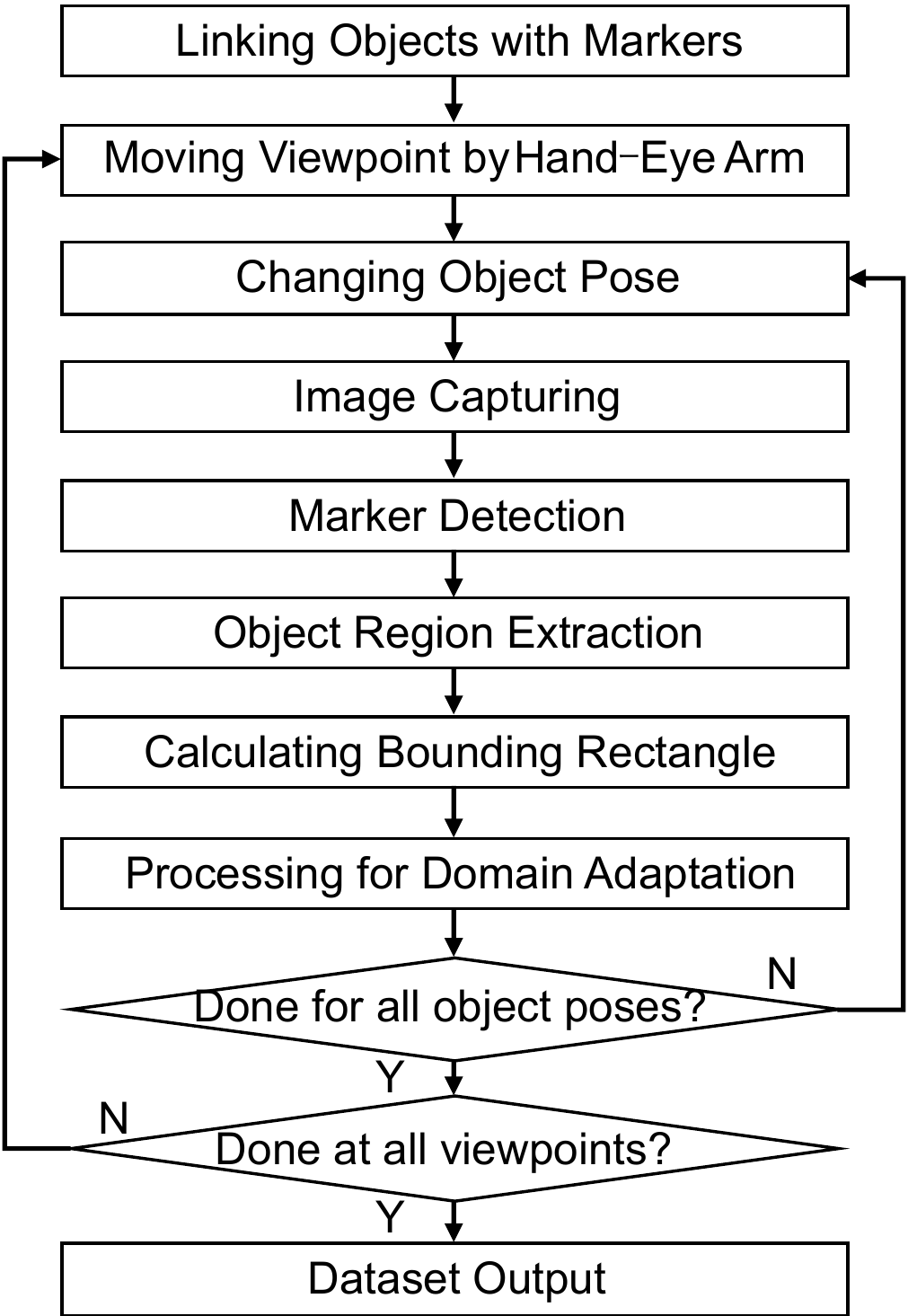}
  \caption{\small Flow of the image dataset collection by the proposed robotic training dataset collection system.}
  \figlab{dataset-collection-flowchart}
\end{figure}

\figref{dataset-collection-flowchart} shows the proposed dataset collection procedure with its automatic annotation method~\cite{RA-L2019}. 
\figref{obj-extract} shows the process for the object region extraction shown in~\figref{dataset-collection-flowchart}.
To extract the region in consideration of the outline blur caused by anti-aliasing, alpha matting is applied to the captured image. We used \textit{large--kernel matting}, a fast method for high quality matting~\cite{AlphaMatting}.
We used a Python library \textit{PyMatting}~\cite{Germer2020} for alpha matting.
Trimap is used for alpha matting and is automatically generated by applying dilation processing to the image that the markers are removed.

The generated approximate object mask is according to the estimated object pose related to the camera.
If coordinate systems for the hand--eye camera, $k$-th visual marker, and the object are $\mathrm{\Sigma}_c$, $\mathrm{\Sigma}_{v_{k}}$, and $\mathrm{\Sigma}_o$, the transformation, ${\bm{M}}_o^c$, from $\mathrm{\Sigma}_c$ to $\mathrm{\Sigma}_o$ shown in~\figref{collection-system} is calculated as
\begin{equation}
{\bm{M}}_o^c={{\bm{M}}_{v_{k}}^c({\bm{r}}_{v_{k}}^c, {\bm{\theta}}_{v_{k}}^c)}{{\bm{M}}_o^{v_{k}}}, \forlab{moc}
\end{equation}
where ${\bm{M}}_{v_{k}}^c$, ${\bm{M}}_{o}^c$, and ${\bm{M}}_{o}^{v_{k}}$ are transformations from $\mathrm{\Sigma}_c$ to $\mathrm{\Sigma}_{v_{k}}$, from $\mathrm{\Sigma}_c$ to $\mathrm{\Sigma}_o$, and from $\mathrm{\Sigma}_{v_{k}}$ to$\mathrm{\Sigma}_o$, respectively.
The translation vector, \textit{${\bm{r}}_{v_{k}}^c$}, and the rotation vector, \textit{${\bm{\theta}}_{v_{k}}^c$}, are estimated from the detected visual markers. 
\begin{figure}[tb]
  \centering
  \includegraphics[width=0.9\linewidth]{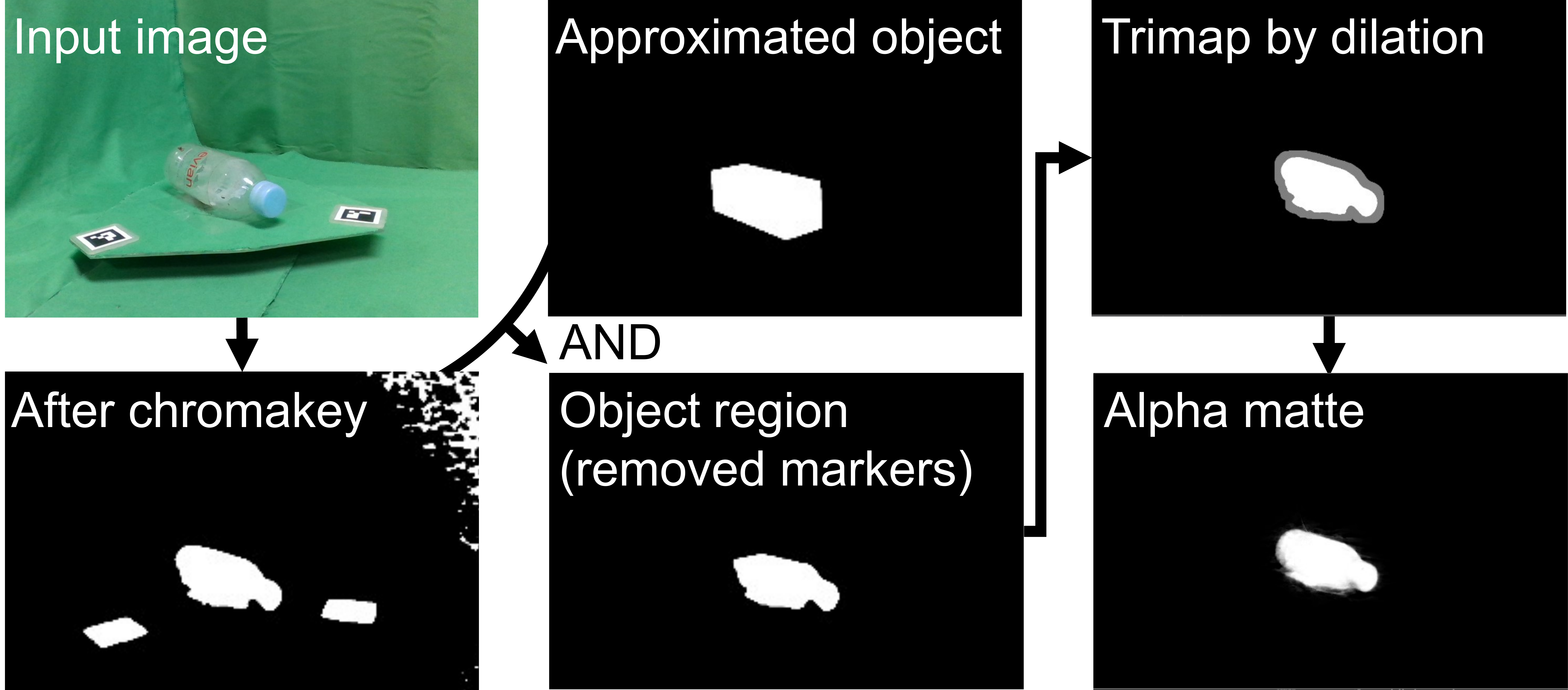}
  \caption{\small Extracted object region (bottom center) by applying AND operation with the image after chromakey (bottom left) and the image showing the approximated object (top center) in the estimated pose based on marker detection, automatically generated trimap (top right), and the generated alpha matte (bottom right) used for alpha matting.}
  \figlab{obj-extract}
\end{figure}
\subsection{Object Image Scaling for Consistency of Geometry}
Object image scaling is applied to the collected images to reduce the differences in appearance caused by the varying distances between the camera and the object. 
To accomplish this, the size of the object placed on the automatic rotating stage is adjusted to be fitted to the size of the object placed on the conveyor in the waste-sorting scene. 

As shown in \figref{transformation}, the visual markers on the marker board in both images are detected. 
For geometric consistency of the dataset images, the size of the object region in the image is adjusted according to the scaling parameter, $k$, estimated as
\begin{gather}
\left( \begin{array}{cc} x'\\ y'\\ \end{array} \right) = \left(\begin{array}{cc} k & 0 \\ 0 & k \\ \end{array} \right) \left( \begin{array}{cc} x\\ y\\ \end{array} \right), \\
k = \frac{d_t}{d_s},
\end{gather}
where $d_s$ and $d_t$ are the distances from the camera coordinate system, $\Sigma_c$, to the marker board coordinate systems, $\Sigma_s$ and $\Sigma_t$, of the source and target images.

\begin{figure}[tb]
  \centering
  \includegraphics[width=0.7\linewidth]{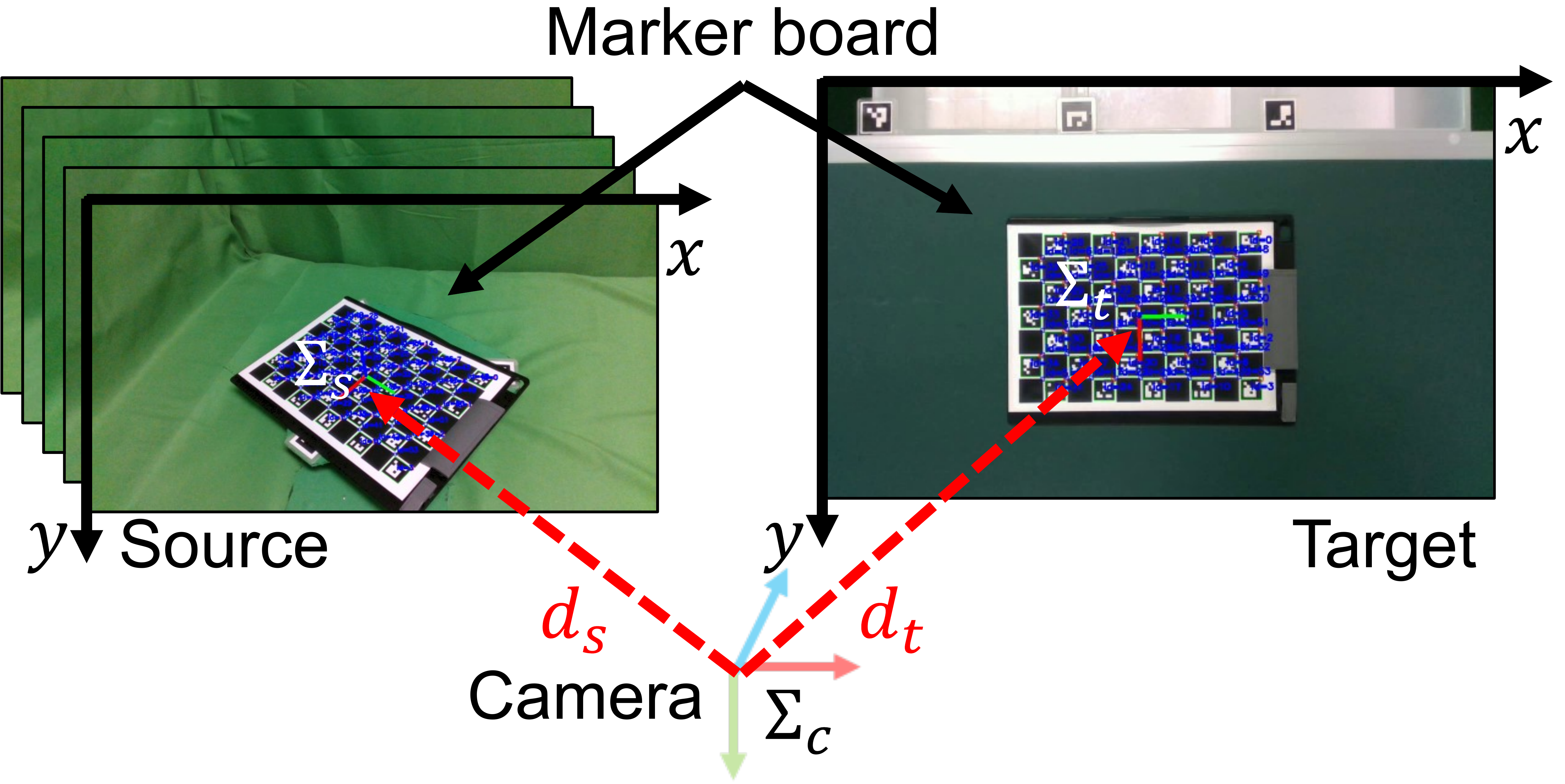}
  \caption{\small Illustration of calculating the scaling parameter, $k$, representing the distances from the camera to the center of the rotating stage used for dataset collection and one point of the conveyor in the waste-sorting scene.}
  \figlab{transformation}
\end{figure}
\subsection{Color Matching and Background Synthesis for Consistency of Illumination}
For the color matching proposed in this study, histograms of pixel values in the RGB color space are calculated from an object-area image captured in the waste-sorting environment, and HM~\cite{DIP2nd} is performed.
The generated image has a distribution similar to the illumination in the waste-sorting environment. Thus, the difference in the illumination is reduced.

The cumulative distribution, $cdf_s(i)~(i = 1,2, .., l)$, of the input image's histogram, $\bm{h}_s$, is matched to the cumulative distribution, $cdf_t(i)$, of target image's histogram, $\bm{h}_t$. 
Each cumulative distribution function (CDF) is calculated as
\begin{equation}
    cdf_{s}(i) = \sum_{j=1}^{i} \frac{h_{s}(j)}{N_s},~~
    cdf_{t}(i) = \sum_{j=1}^{i} \frac{h_{t}(j)}{N_t},
\end{equation}
where $l$ is the number of bins in the histogram, and $N_t$ and $N_s$ are the number of pixels in each image.

To extract the boundary between the object and the background, using the automatically generated trimap, we apply alpha blending~\cite{AlgApp} to the image at the time of image collection to combine it with the background image captured in the waste-sorting environment. 
Then, we apply HM to the image of only the area within the bounding box of the object. 

Images used for applying HM to the image of the plastic bottle are shown in \figref{color-match}. 
The leftmost image shows the source image, the image to the right of the source image is a target image as the destination, the image to the right of the target image shows a result of the HM, and the rightmost image shows the image after EQ.
We use \textit{Contrast Limited Adaptive Histogram Equalization (CLAHE)}~\cite{CLAHE1994} to smooth jaggy histogram distributions by the EQ.
Finally, background-synthesized and histogram-matched images are used to train the waste detector.
\begin{figure}[tb]
  \centering
  \includegraphics[width=0.96\linewidth]{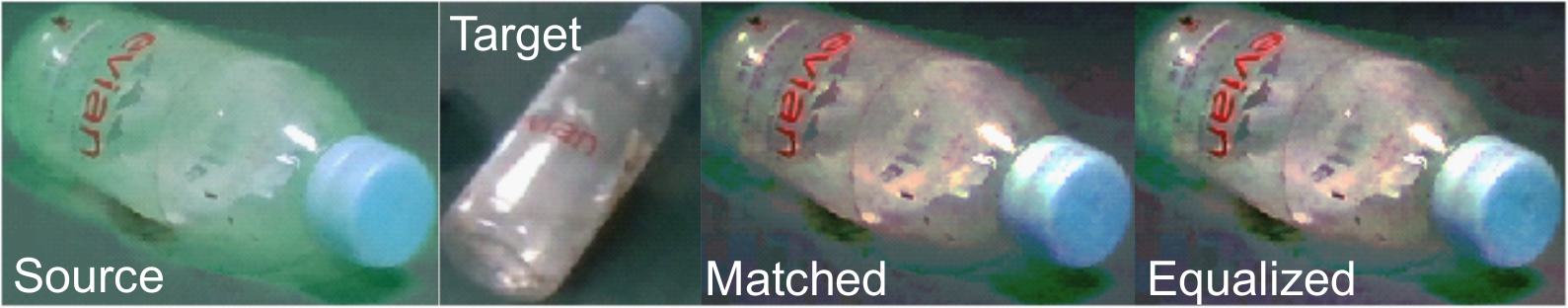}
  \caption{\small HM applied to a plastic bottle image. ``Source'' and ``Target'' indicate the input image and the image with the target histogram to match. ``Matched'' and ``Equalized'' are the images after application of HM and after application of the EQ of Matched, respectively.}
  \figlab{color-match}
\end{figure}

\section{Agile Handling of Conveyed Objects}
\subsection{Two Types of Sorting Manipulation} \label{subsec:two-manip}
In this study, one sorting task is designated to move a waste item from the conveyor to an adjacent recycling box.
Regarding the waste-sorting robot, quickness is required alongside sorting accuracy. 
Therefore, two types of sorting manipulation are performed according to the desired waste detection results. 

As shown in \figref{manipulation}, the two types are (1) manipulation by picking and releasing and (2) manipulation by pushing and dropping. 
A gripper with one degree of freedom can perform these manipulations.

In manipulation (1), the object is grasped by the five fingers of the soft gripper so that the estimated point near the object center (\textit{virtual CoM}) becomes to the grasping center. 
The gripper pose is adjusted to enable grasping along the straight line on the estimated object silhouette passing through the virtual CoM, which is illustrated in \figref{manipulation}(a). 
Then, the robot arm trajectory is planned and generated so that it approaches the target object and departs from it in its fixed grasping pose.

In manipulation (2), the soft gripper pushes the object around the virtual CoM using a straight-line trajectory and drops the object into the target recycling bin (\figref{manipulation}(b)). 
The trajectory of a robot arm is generated to push the object in a direction that connects the virtual CoM to the front center of the recycling bin.

In contrast to pick-and-release, push-and-drop does not require grasping. 
The average time in the 10 trials to finish the push-and-drop operation was 3.3~s, although the time in the case of the pick-and-release operation took 5.2~s.
Thus, using the push-and-drop operation as much as possible shortens the combined manipulation time.

\subsection{Selective Execution and Implementation}
\algoref{sorting} describes the entire algorithm used to select a feasible manipulation from the two available types. 
The algorithm is based on the following policy considering the time constraints of feasible manipulation to handle the waste items conveyed.
\begin{enumerate}
  \item We adapt a first-in-first-out strategy to determine how to manipulate the frontmost waste item on the conveyor.
  \item Push-and-drop is primary performed if possible because of its quickness.
  \item If both types of manipulation are determined to be infeasible based on the time constraints, the target waste item is ignored (shown as ``\textbf{continue}'' in \algoref{sorting}).
\end{enumerate}

The positions indicated by the parameters are drawn in \figref{explain-params}.
We define the width and height of the object silhouette in the image as $s_x$ and $s_y$, respectively. We define the x- and y-axial distances from the center of the target object's silhouette to the recycling bin's center line as $l_{bx}$ and $l_{by}$, respectively.
These are calculated from the object's silhouette mask image and the results of a detected marker attached to the recycling bins. 
$l_{e}$ is the x-axial distance from the object to the image right end.

$t_{pd}$ and $t_{pp}$ are the time variables representing the times required for push-and-drop and pick-and-release manipulation, respectively. 
These are determined in preliminary experiments to measure the manipulation time in all points on the conveyor and the premeasured gripper open--close time. 

$v_{pd}$ and $v_{c}$ are the speed variables for the push-and-drop manipulation and transportation of conveyor.
These are preset parameters (\ie~the speed of the push-and-drop manipulation and the transportation speed of the conveyor are constant for the waste-sorting).

Here, we consider following three time constraints to select the manipulation type in~\algoref{sorting}.
\begin{enumerate}
  \item The inequality, $(s_{x}/2)/v_{c} > l_{by}/v_{pd}$, holds in the cases where target waste item is far from bins and too small to push. This indicates that it is impossible to execute the push-and-drop operation.
  \item The inequality, $t_{pp} < l_{e}/v_{c}$, holds in the state that the target waste item cannot be manipulated in the pick-and-release manipulation time.
  This indicates that it is too late to start the pick-and-release.
  \item The inequality, $t_{pd} < l_{bx}/v_{c}$, holds in the state that the target waste item is conveyed to a position where the robot cannot push it into the target bin. This indicates that it is too late to start the push-and-drop.
\end{enumerate}
\begin{figure}[tb]
  \centering
  \subcaptionbox{\small{Pick-and-release}}{\includegraphics[width=0.4\linewidth]{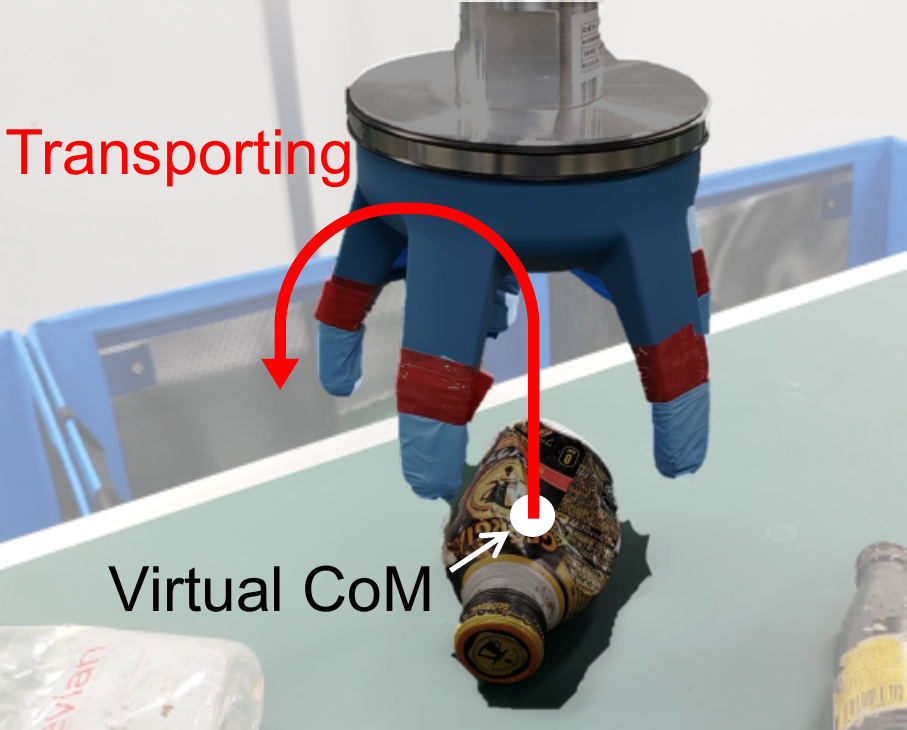}\hspace{0mm}}
  \subcaptionbox{\small{Push-and-drop}}{\includegraphics[width=0.4\linewidth]{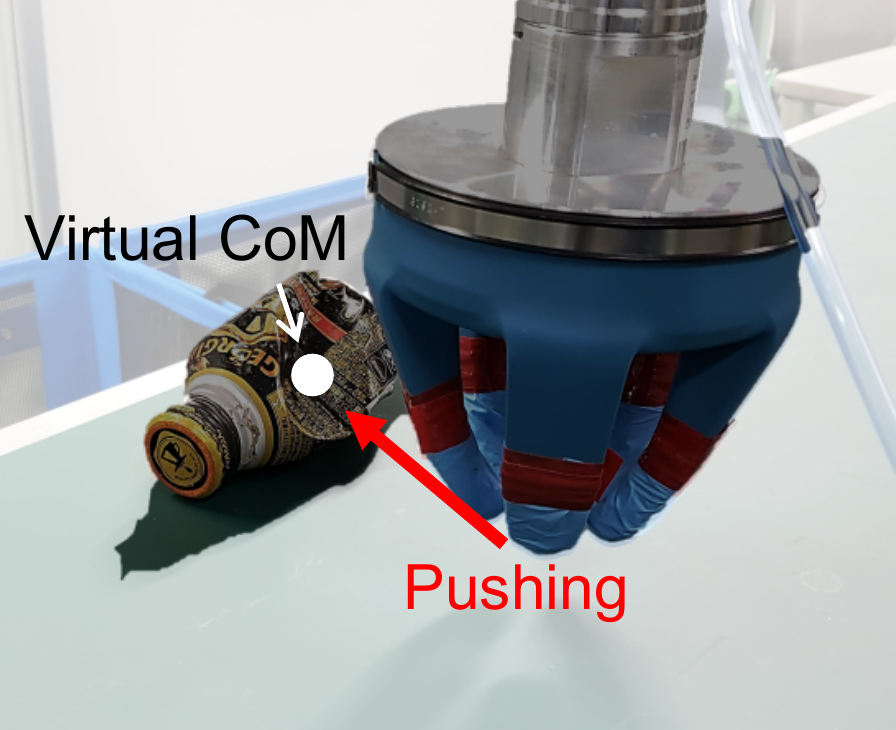}\hspace{0mm}}
  \vspace{1mm}
  \caption{\small Illustration of key scenes in the two proposed types of manipulation (\ie~(a) pick-and-release and (b) push-and-drop) to sort the waste (\ie~aluminum can, glass bottle, and plastic bottle) on a conveyor belt.}
  \figlab{manipulation}
\end{figure}
\begin{figure}[tb]
  \centering
  \includegraphics[width=0.65\linewidth]{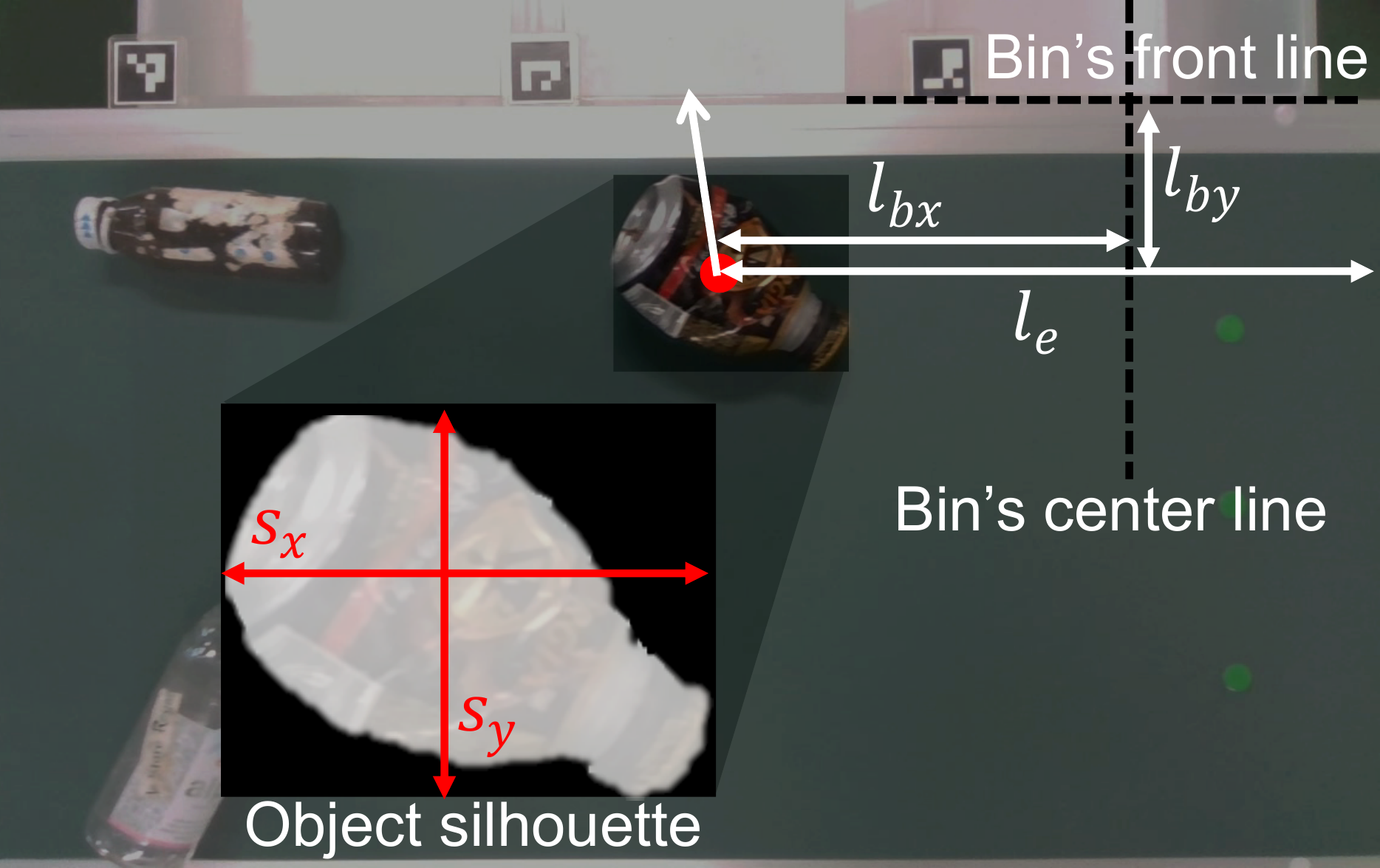}
  \caption{\small Parameterization for the sorting manipulation selection algorithm.}
  \figlab{explain-params}
 \end{figure}
\alglanguage{pseudocode}
\begin{algorithm}[t]
\caption{Sorting Manipulation Selection}
\algolab{sorting}
\begin{algorithmic}[1]
\renewcommand{\algorithmicrequire}{\textbf{Input:}}
\renewcommand{\algorithmicensure}{\textbf{Output:}}
\Require An image of one place on the conveyor
\Procedure{SELECT-MANIPULATION-TYPE}{}
\State Recognize waste items and markers in the image
\State Calculate $s_x$ and $s_y$ from the object silhouette mask
\State Calculate $l_{bx}$, $l_{by}$ and $l_e$ from the recognition results
\If {$(s_{x}/2)/v_{c} > l_{by}/v_{pd}$}
    \If {$t_{pp} < l_{e}/v_{c}$}
        \State \textbf{continue}
    \EndIf
    \State Execute pick-and-release on robot
    \State \textbf{continue}
\EndIf
\If {$t_{pd} < l_{bx}/v_{c}$}
    \State \textbf{continue}
\EndIf
\State Execute push-and-drop on robot
\EndProcedure
\end{algorithmic}
\end{algorithm}

Detected objects are assigned silhouettes extracted using the input-depth image. 
Using the known distance from the RGBD sensor to the conveyor, we create the silhouette mask as the object regions on the conveyor that are closer to the camera than the conveyor.
The centroids of the silhouette pixel areas are estimated for each object as the virtual CoM. 

All parameters are estimated from RGB and depth images of one frame to maintain fast computations for the waste detector. 
The virtual CoM from the 2.5-dimensional RGBD image reflect an ill-posed problem. 
We assume that an object's shape can be approximated as a solid revolution with a uniformly distributed mass. 
The underlying assumption enables us to estimate the virtual CoM using the object silhouette extracted from the RGB and depth image.
The virtual CoM is calculated as the centroid of a grayscale image.

Using the estimated common parameters, the unique parameters (\ie~grasp position and pushing direction) are calculated based on the methods mentioned in Section \ref{subsec:two-manip}. 
The arm motions for picking, releasing, pushing, and dropping and their connecting trajectories are planned and generated using \textit{MoveIt!}~\cite{ChittaRAM2012}.

\section{Experiments}
\begin{figure*}[tb]
  \centering
  \subcaptionbox{\small{Aluminum can}}{\includegraphics[width=0.98\linewidth]{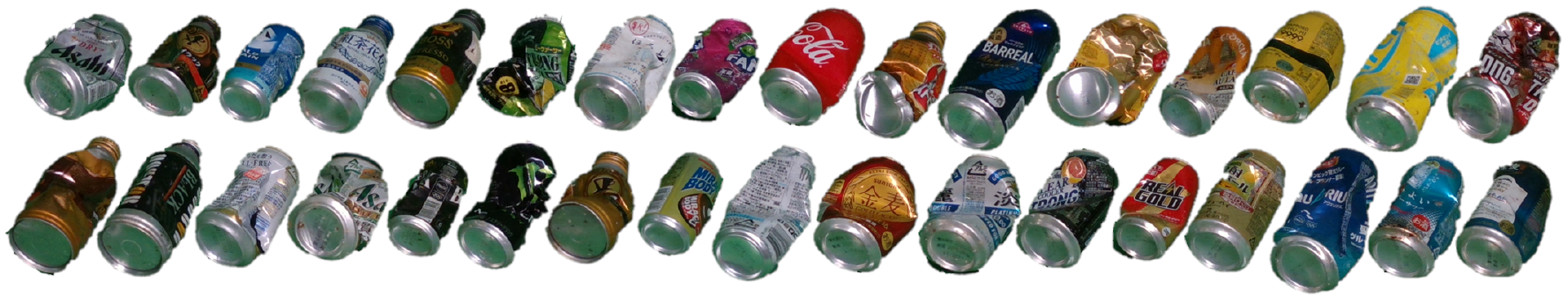}\vspace{-2mm}\hspace{0mm}}
  \subcaptionbox{\small{Glass bottle}}{\includegraphics[width=0.98\linewidth]{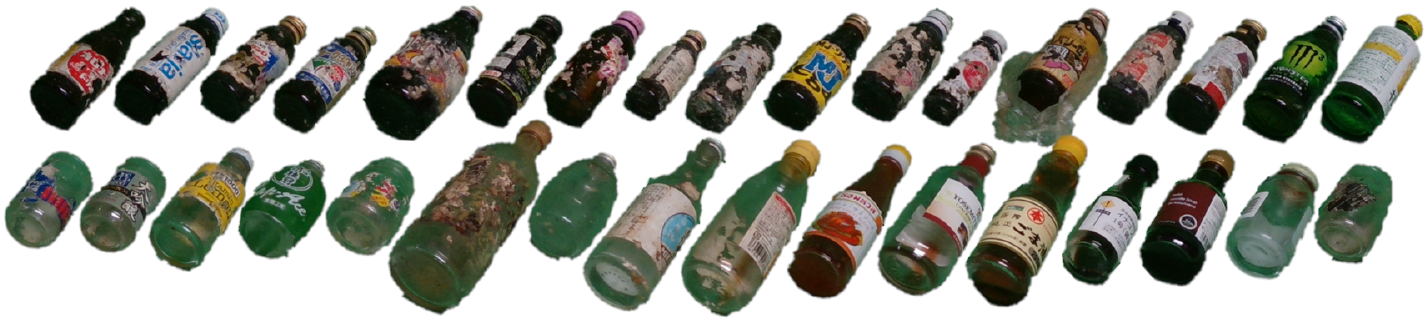}\hspace{0mm}}
  \subcaptionbox{\small{Plastic bottle}}{\includegraphics[width=0.98\linewidth]{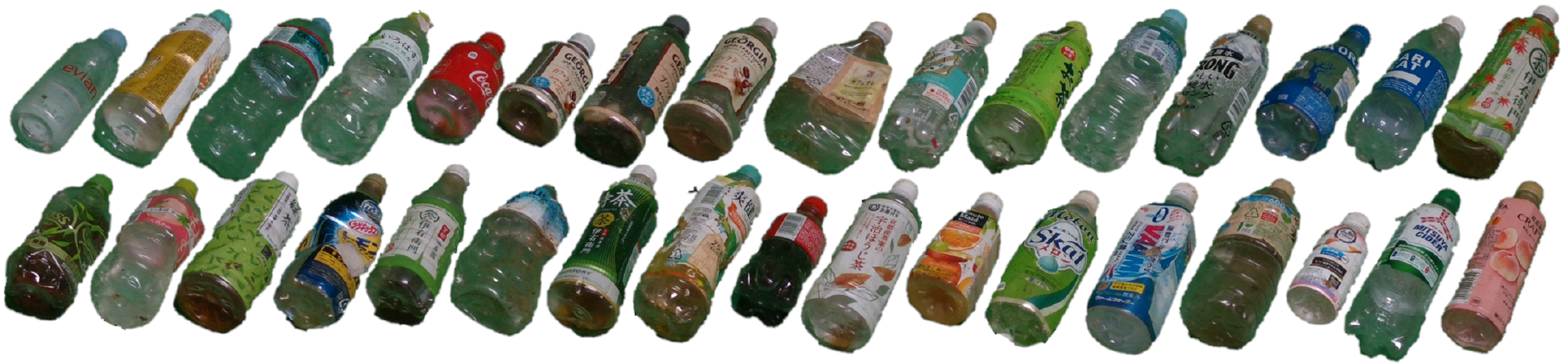}\vspace{-2mm}\hspace{0mm}}
  \caption{\small The waste samples of (a) aluminum cans; (b) glass bottles; and (c) plastic bottles used in the experiments.}
  \figlab{target-objects}
\end{figure*}
\begin{table*}[tb]
    \caption{\small Average time to collect 100 image datasets. Automatic or manual is shown next to the time measured.}
    \centering
    \begin{tabular}{cp{15mm}p{25mm}p{15mm}p{25mm}p{15mm}p{25mm}} \toprule
        & \multicolumn{6}{c}{{Type of automatic dataset collection}} \\ \cmidrule(r){2-7}
        & \multicolumn{2}{c}{{Single marker~\cite{RA-L2019}}} & \multicolumn{2}{c}{{Multiple markers~\cite{AdvancedRobotics2019}}} & \multicolumn{2}{c}{{Proposed}} \\ 
        {Necessary process} & \hfil{Time [s]}\hfil & \hfil{Automatic / Manual}\hfil & \hfil{Time [s]}\hfil & \hfil{Automatic / Manual}\hfil & \hfil{Time [s]}\hfil & \hfil{Automatic / Manual}\hfil \rule[0pt]{0pt}{8pt} \\ \midrule
        {Object replacement} & \hfil{\multirow{2}{*}{900}}\hfil & \hfil{Manual}\hfil & \hfil{\multirow{2}{*}{5232}}\hfil & \hfil{Manual}\hfil & \hfil{2.05}\hfil & \hfil{Manual}\hfil \rule[0pt]{0pt}{8pt} \\
        {Image acquisition} &  & \hfil{Manual}\hfil &  & \hfil{Manual}\hfil & \hfil{10.2}\hfil & \hfil{Automatic}\hfil \rule[0pt]{0pt}{8pt} \\
        {Annotation} & \hfil{444}\hfil & \hfil{Automatic}\hfil & \hfil{-}\hfil & \hfil{Automatic}\hfil & \hfil{-}\hfil & \hfil{Automatic}\hfil \rule[0pt]{0pt}{8pt} \\ \midrule
        {Total} & \hfil{1344}\hfil & \hfil{-}\hfil & \hfil{5232}\hfil & \hfil{-}\hfil & \hfil{12.3}\hfil & \hfil{-}\hfil \rule[0pt]{0pt}{8pt} \\ \bottomrule 
    \end{tabular}
    \tablab{collection-time}
\end{table*}
\begin{figure*}[tb]
  \centering
  \subcaptionbox{\small{Bird's-eye view}}{\includegraphics[width=0.325\linewidth]{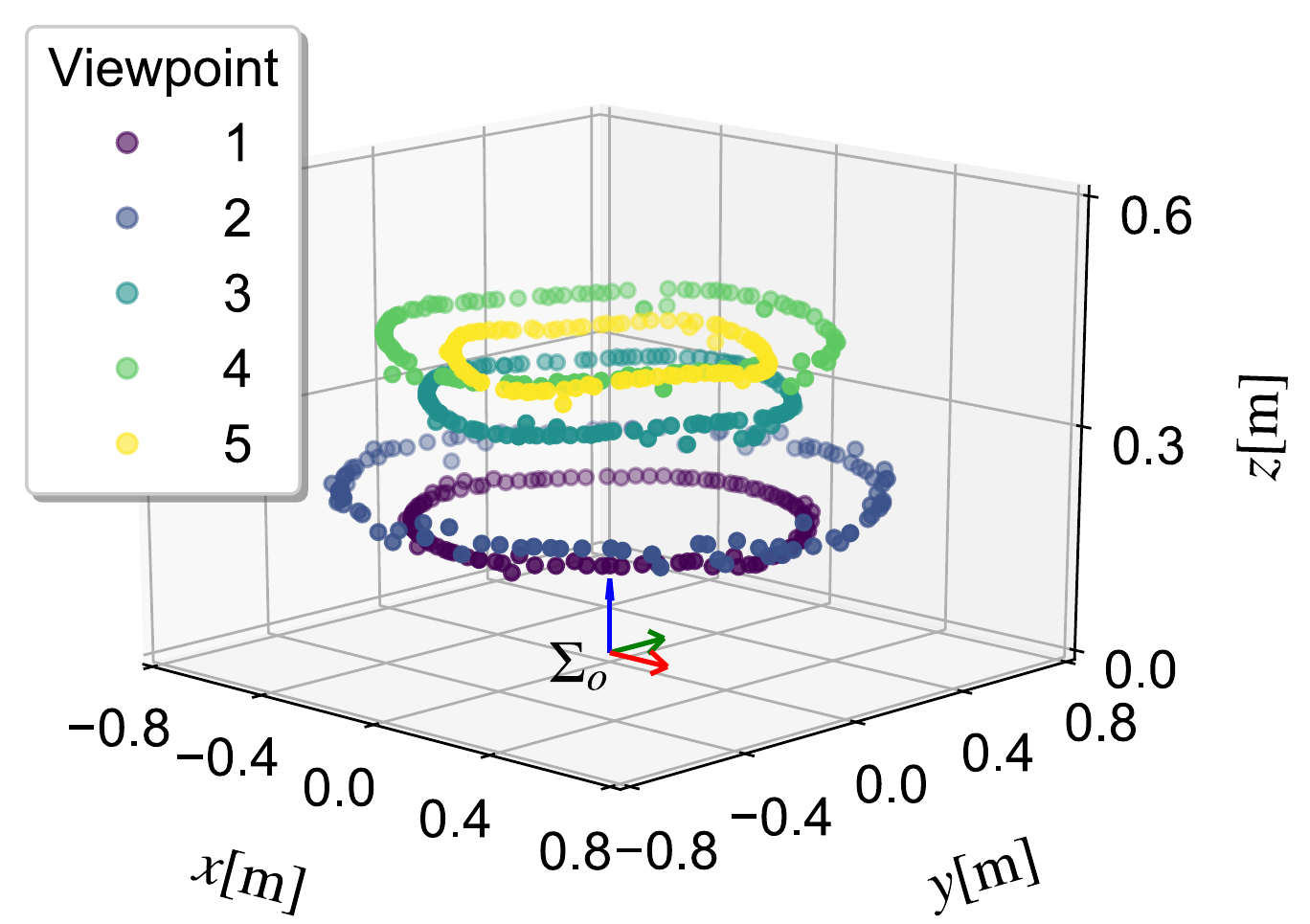}}
  \subcaptionbox{\small{Top view}}{\includegraphics[width=0.325\linewidth]{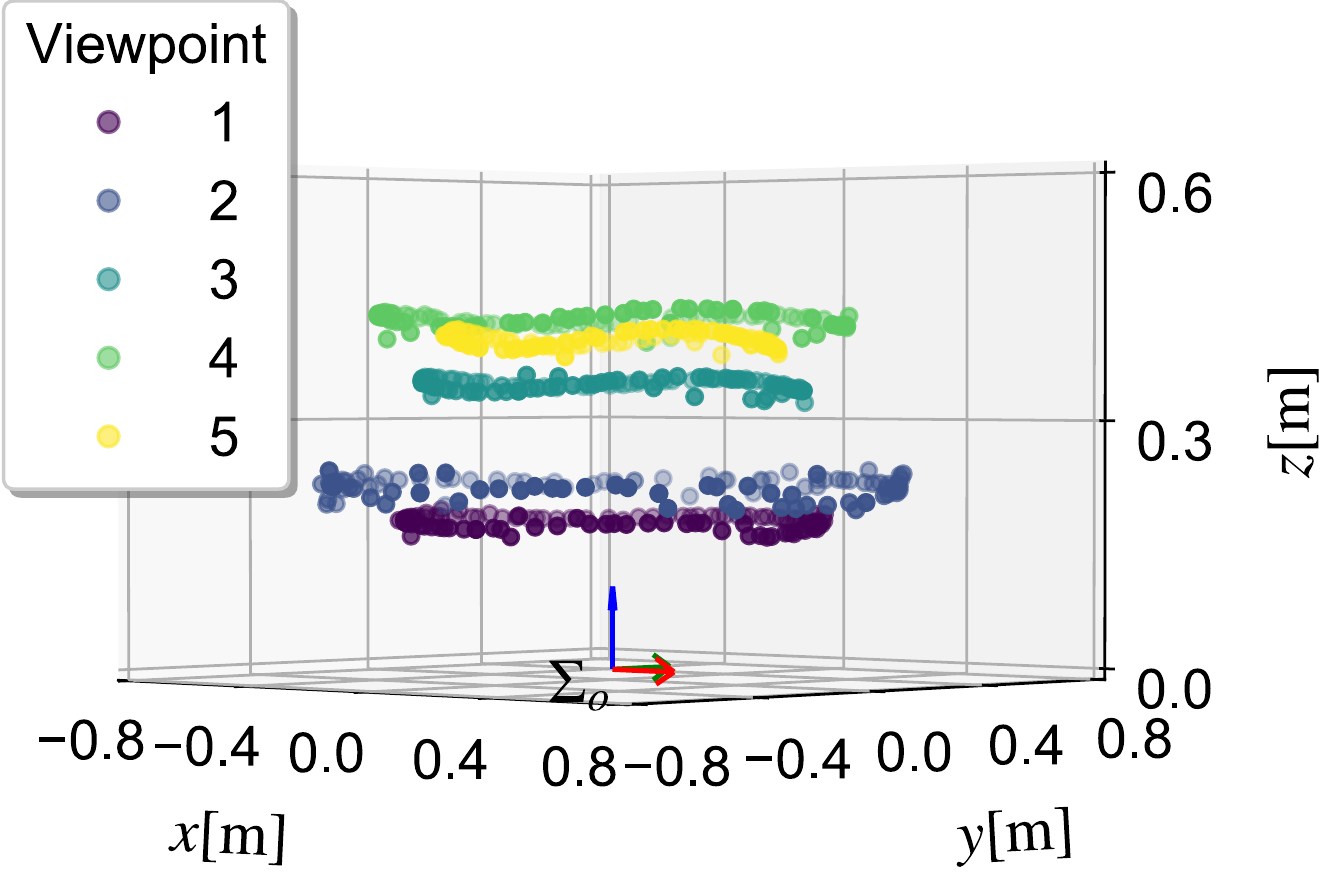}}
  \subcaptionbox{\small{Side view}}{\includegraphics[width=0.325\linewidth]{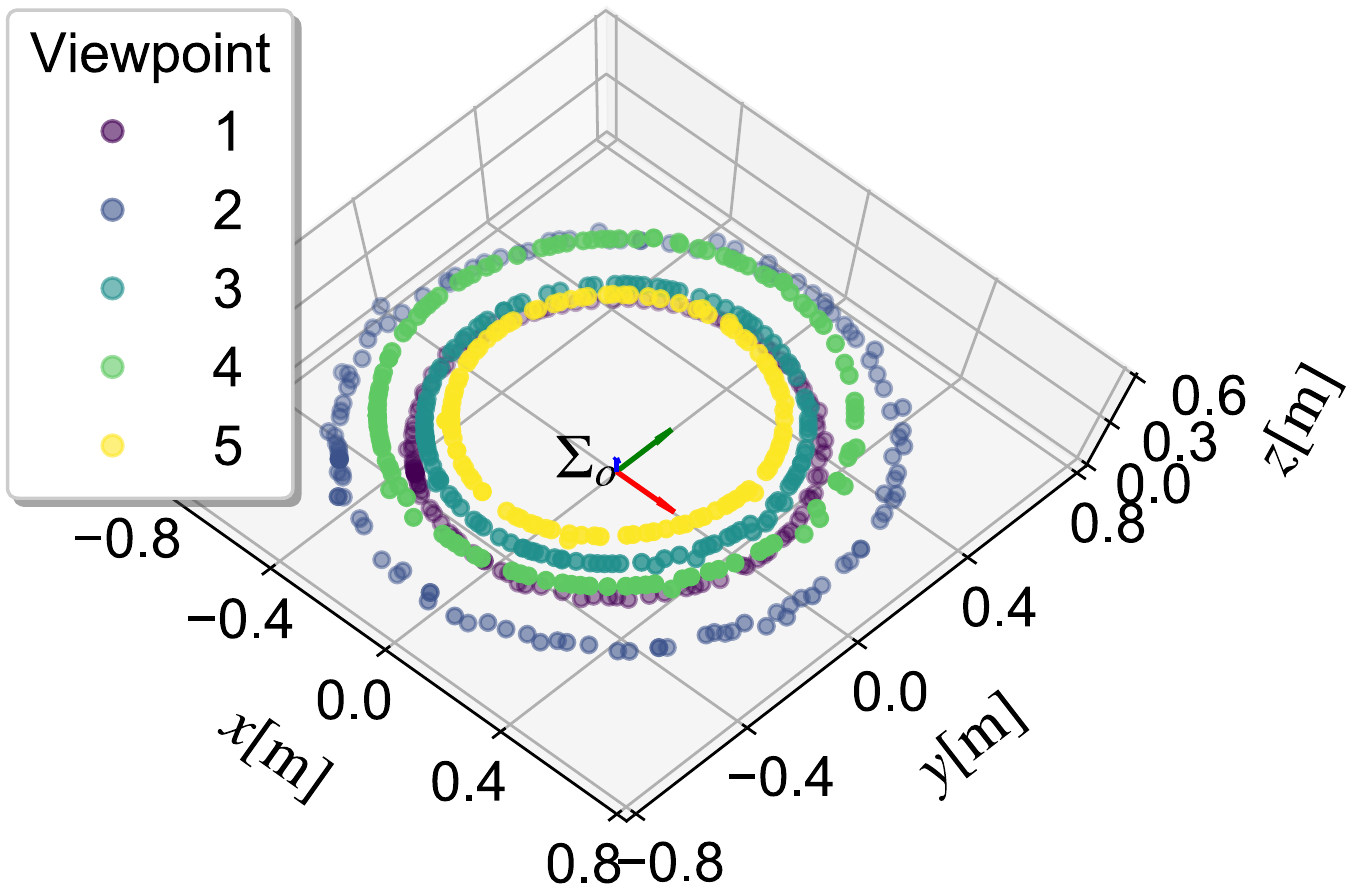}}
  \caption{\small Variations of viewpoints taken by the proposed robotic training dataset collection system. $\Sigma_o$ shows the object coordinate system shown in~\figref{collection-system}. Viewpoint IDs from 1 to 5 represent the five viewpoint patterns adjusted by changing the joint pose of the small robot arm.}
  \figlab{variation}
\end{figure*}
\subsection{Outline of Experiments}
First, to evaluate the quickness of the proposed robotic training dataset collection system, we compare the collection time by the proposed dataset generation with the collection time by the manual dataset generation (Section~\ref{collection_time}).
Furthermore, we show the accuracy of the annotation results (Section~\ref{annotation_accuracy}).

Second, we show the similarity of the images applied adaptation methods with those captured from the real scene (Section~\ref{similarity}).
To evaluate the performance of the waste detector trained with the image dataset that applied the proposed adaptation method having the highest similarity, we show the detection results of the sorting-target waste item by the detector (Section~\ref{detection_result}).

Third, to evaluate the feasibility of the proposed manipulation methods, we discuss the success rate of the sorting manipulation and the average time required by each manipulation method (\ie~pick-and-release and push-and-drop) (Section~\ref{feasibility}).

\subsection{Experimental Setup}
As shown in \figref{collection-system}, we used \textit{COBOTTA} (DENSO WAVE INCORPORATED) with \textit{RealSense D435} (Intel Inc.) as the small hand--eye robot and used \textit{OSMS-60YAW} (SIGMAKOKI CO.,LTD.) as the rotating stage.
We used \textit{ArUco}, an AR library~\cite{ArucoMain,ArucoAdditional} to detect AR markers for registering the object pose of each object image collected using the proposed robotic training dataset collection system. 
This object poses were used to generate an approximate object mask.
ArUco was used to specify the positions of the recycling bins in the waste-sorting experiments.

An evaluation experiment of the waste-sorting was performed using the robotic waste-sorting system shown in \figref{system}.
In this paper, we experimented with the minimum configuration of one camera and one manipulator.

We used a robot arm, \textit{LBR iiwa 14 R820} (KUKA), and a soft gripper, \textit{SOFTmatics} (Nitta Corporation), whose five fingers were covered with a soft material to handle the many sharp objects present in a recycling facility. 
We used an RGBD camera employing active infrared stereo, the same RealSense camera as the camera used in the dataset collection. The camera can measure depth information with high sensitivity, even for translucent objects and those having complex shapes and opacity, which are common in container and packaging waste.
The target waste samples contained 33 different aluminum cans, 33 glass bottles, and 33 plastic bottles, as shown in \figref{target-objects}. 
The target objects were sampled from the waste samples in a recycling factory for industrial waste items.

\subsection{Image Dataset Collection Time} \label{collection_time}
To demonstrate the effectivity of the robotic training dataset collection system compared with the collection methods previously proposed in~\cite{RA-L2019,AdvancedRobotics2019}, this section describes the results of the comparison of times needed to collect image datasets.

\tabref{collection-time} shows the average time needed to collect 100 images and the method (automatic or manual) for three processes: object replacement, image acquisition, and annotation. 

In the first proposed method using a single marker~\cite{RA-L2019}, an object with a marker attached directly was actually used in a real-work environment.
In this method, humans manually change the object types and the poses of the objects. Therefore, it took a relatively long time of 900 seconds.   
In addition, the annotation was automatically performed by image processing after all the image capturing was completed. 
As the result, it was 444 seconds for 100 images.
In the extended method using multiple markers~\cite{AdvancedRobotics2019}, object replacement and image acquisition were performed manually as in the single marker method. 
Combined with the time required for these manual operations and the time required for automatic annotation that was being processed in parallel, it took the longest time of 5232 seconds.

The proposed dataset collection was completed in 12.3~s on average for 100 images. The results indicate that the time required for collecting the training set was incredibly shortened compared with the other methods.
The viewpoints taken by the proposed robotic training dataset collection system are widely scattered as shown in~\figref{variation}, suggesting that a dataset having large variations can be collected in a short period.

The total time required to collect the training set comprising 59,400 (120 object-orientation patterns $\times$ 5 viewpoint patterns $\times$ 99 objects) images captured with a green screen was about 111 min. Such a short collection time enables us to easily increase the number of training sets when the target waste increases or changes.

\subsection{Quantitative Evaluation of Annotations} \label{annotation_accuracy}
To evaluate the annotation results, the automatically object-extracted image is compared with the manually annotated image, as shown in~\figref{anno-result}.
Using a manual annotation tool named \textit{labelme}\footnote[1]{https://github.com/wkentaro/labelme} and by clicking several points on the object contour in images, the images are annotated by humans for evaluation.

Based on true-positive (TP), false-positive (FP), and false-negative (FN) results, as shown in~\figref{anno-result}, we calculated the intersection over union (IoU), precision, recall, and F-score~\cite{Wang2020} as
\begin{gather} \label{metric}
\mathrm{IoU} = \frac{\mathrm{TP}}{\mathrm{TP} + \mathrm{FP} + \mathrm{FN}}, \\ \hspace{2mm}
\textrm{F-score} = \frac{2 \times \mathrm{Precision} \times \mathrm{Recall}}{\mathrm{Precision} + \mathrm{Recall}}, \\ \hspace{2mm}
\mathrm{Precision} = \frac{\mathrm{TP}}{\mathrm{TP} + \mathrm{FP}}, \\ \vspace{2mm}
\mathrm{Recall} = \frac{\mathrm{TP}}{\mathrm{TP} + \mathrm{FN}}.
\end{gather}
\tabref{data-seg-res} shows the results of the object region extraction in the training set. 

In all trials and categories, the mean values of precision rated around 70\%.
The mean values of recall were rated higher than 95\% and with smaller standard deviations than those of precision.
These results suggest that there were some false predictions.
However, there were few missed pixels in the ground truth.
As a result, the calculation provides a low IoU with a mean of F-score of 80\%. 

\subsection{Effect of Reducing Differences from Waste-sorting Scene} \label{similarity}
In this section, we discuss the effect of the proposed method of reducing the differences from the waste-sorting scene. 
To evaluate the performance of the proposed color adjustment, we compare it with two other methods. 

The first unifies color reproducibility by applying \textit{color correction} (CC) using \textit{ColorChecker Passport Photo} (X-Rite, Inc.), which has a panel of 24 industry-standard color-reference chips. 
The CC in this study is based on a color-transfer method that can adjust the colors in an image to match a target-image color profile~\cite{PlantCV}. 
The goal is to create a transform so that, when it is applied to the values of every pixel in a source image (the left of \figref{checker}), it returns values mapped to a target image (the right of \figref{checker}) profile~\cite{ColorHomography}. 

The other is an easy-to-use image-rendering SC method~\cite{PIE2003} used in the fields of computer graphics~\cite{Kakuta2007} and computer vision~\cite{Mukaigawa2001,Sato2005,Yokoya2015}. 
SC was once used to create a photomontage by pasting an image region onto a new background using Poisson image editing~\cite{PIE2003}. 
\figref{both-proc} shows the results of CC, SC, and HM.
The parameters needed in the methods described in this section are organized in the \tabref{required-params}.
\begin{figure}[tb]
  \centering
  \subcaptionbox{\small{Clicked points in annotation tool}}{\includegraphics[width=0.665\linewidth]{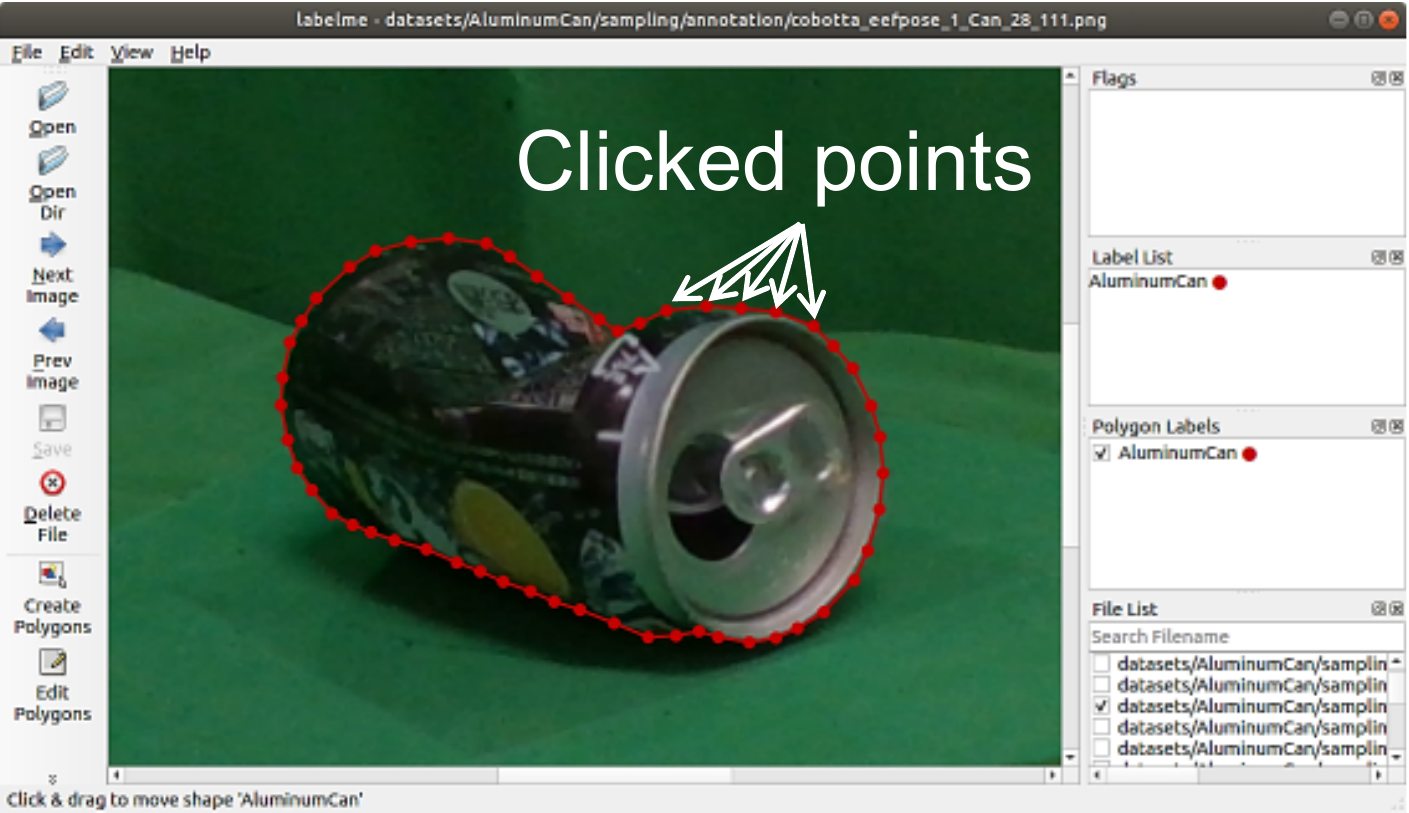}}
  \subcaptionbox{\small{Evaluation}}{\includegraphics[width=0.3\linewidth]{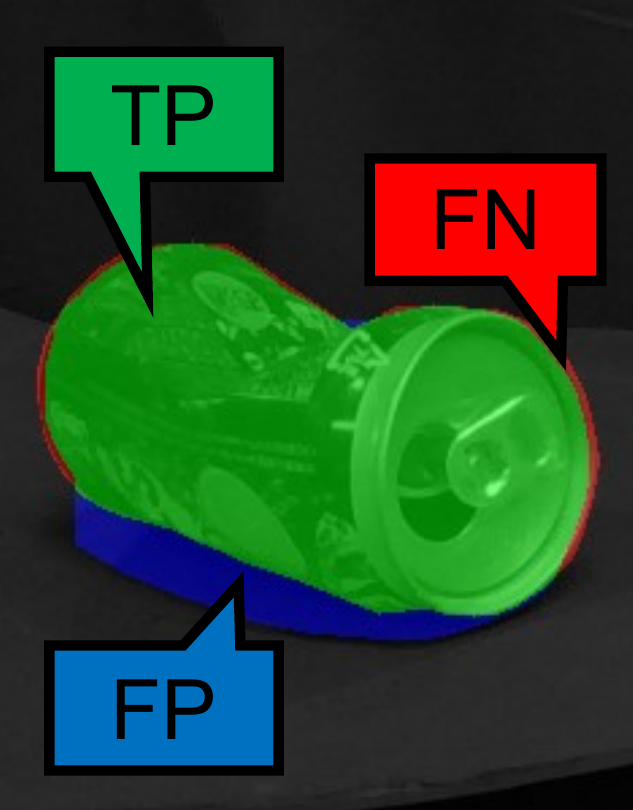}}
  \caption{\small Visualization of manual annotations needed to generate ground truth to evaluate the proposed automatic object region extraction. Left image shows the window of the annotation tool (labelme) and several annotated image points. Right image shows the parameterization of the evaluation results of the automatic object region extraction.}
  \figlab{anno-result}
\end{figure}
\begin{table}[tb]
    \centering
    \caption{\small Results of our object region extraction in our automatic dataset generation. Each element shows mean$\pm$standard deviation of IoU [\%], Precision [\%], Recall [\%], and F-score [\%].
    The mean values are calculated from randomly selected 33 images of each object category in the three categories.}
    \begin{tabular}{p{17mm}p{11mm}p{11mm}p{11mm}p{11mm}} \toprule
        & \multicolumn{4}{c}{{Metric [\%]}} \\ \cmidrule(r){2-5}
        \hfil{Object}\hfil & \hfil{IoU}\hfil & \hfil{Precision}\hfil & \hfil{Recall}\hfil & {F-score} \\ \midrule
        \hfil{Aluminum can}\hfil & \hfil{71$\pm$16}\hfil & \hfil{71$\pm$17}\hfil & \hfil{98$\pm$1.7}\hfil & \hfil{81$\pm$11}\hfil \rule[0pt]{0pt}{8pt} \\
        \hfil{Glass bottle}\hfil & \hfil{67$\pm$17} & \hfil{69$\pm$18}\hfil & \hfil{96$\pm$4.4}\hfil & \hfil{79$\pm$13}\hfil \rule[0pt]{0pt}{8pt} \\ 
        \hfil{Plastic bottle}\hfil & \hfil{77$\pm$14} & \hfil{78$\pm$15}\hfil & \hfil{97$\pm$2.2}\hfil & \hfil{86$\pm$9.8}\hfil \rule[0pt]{0pt}{8pt} \\ \bottomrule
    \end{tabular}
    \tablab{data-seg-res}
\end{table}
\begin{figure}[tb]
  \centering
  \subcaptionbox{\small{Color correction}}{\includegraphics[width=0.32\linewidth]{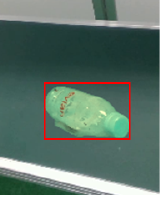}}
  \subcaptionbox{\small{Seamless cloning}}{\includegraphics[width=0.32\linewidth]{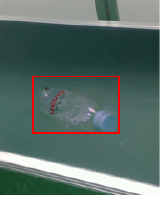}}
  \subcaptionbox{\small{Histogram matching}}{\includegraphics[width=0.32\linewidth]{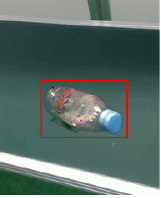}}
  \caption{\small Comparison of appearances of synthesized images: (a) synthesized images with CC applied; (b) SC applied; and (c) HM applied.}
  \figlab{both-proc}
\end{figure}
\begin{figure*}[tb]
  \centering
  \includegraphics[width=0.95\linewidth]{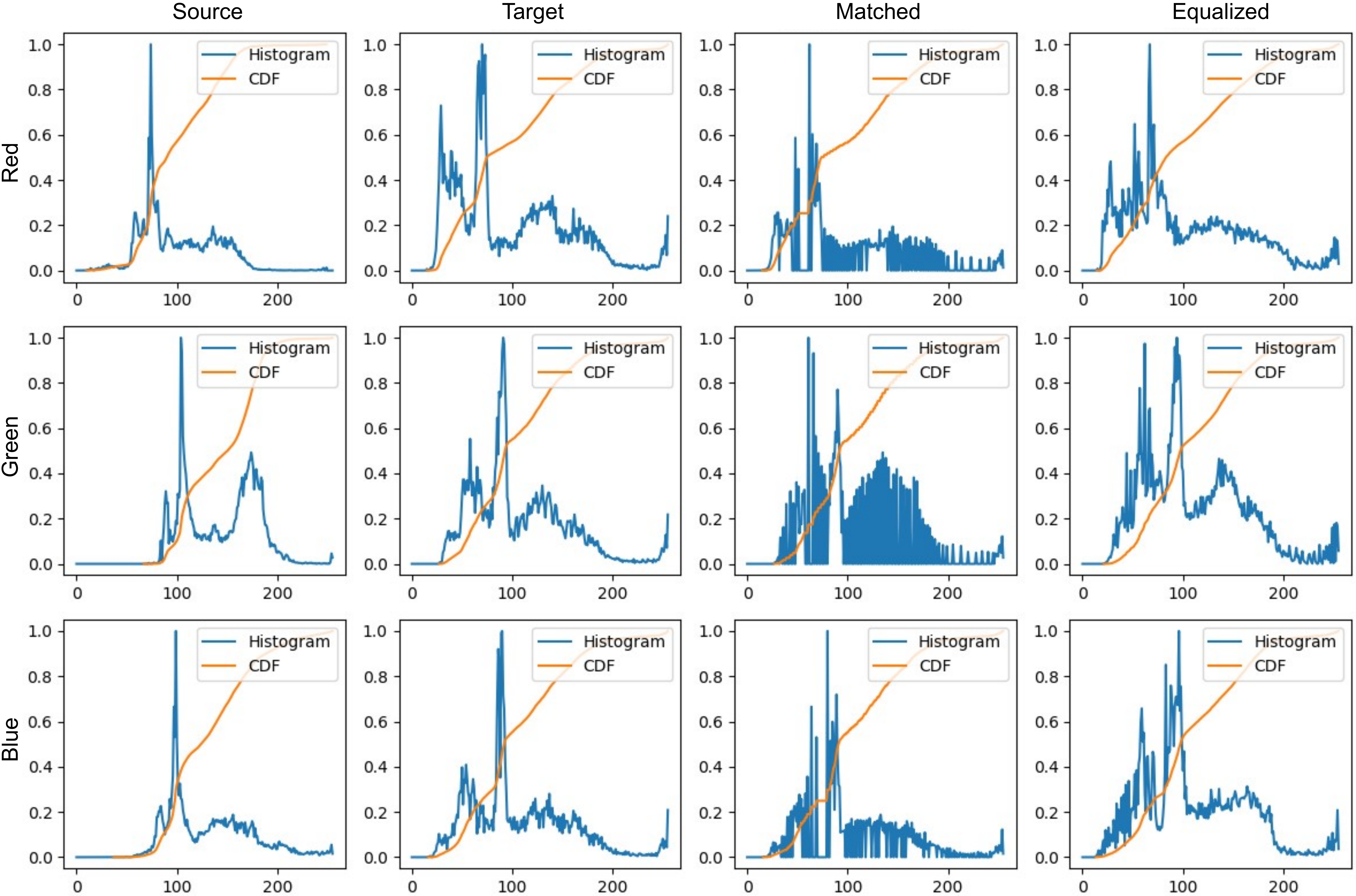}
  \caption{\small RGB histograms and CDFs of the image applied with HM. The graphs are histograms of the RGB color space of the four images on~\figref{color-match}. The title names correspond to the names displayed in each image shown in~\figref{color-match}.}
  \figlab{color-match-result}
\end{figure*}
\begin{figure}[tb]
  \centering
  \includegraphics[width=0.92\linewidth]{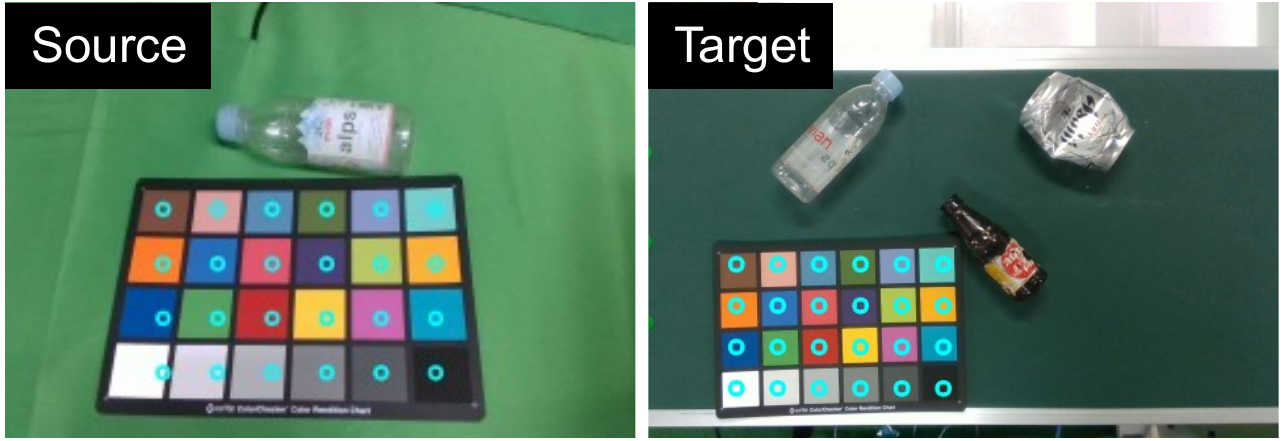}
  \caption{\small Images used for estimating the color homography transformation matrix for CC.}
  \figlab{checker}
\end{figure}
\begin{figure*}[tb]
  \centering
  \subcaptionbox{\small{\vspace{1mm}Image sequence recorded in the waste-sorting environment}}{\includegraphics[width=\linewidth]{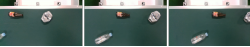}}
  \subcaptionbox{\small{Visualization of detection results of our waste detector}}{\includegraphics[width=\linewidth]{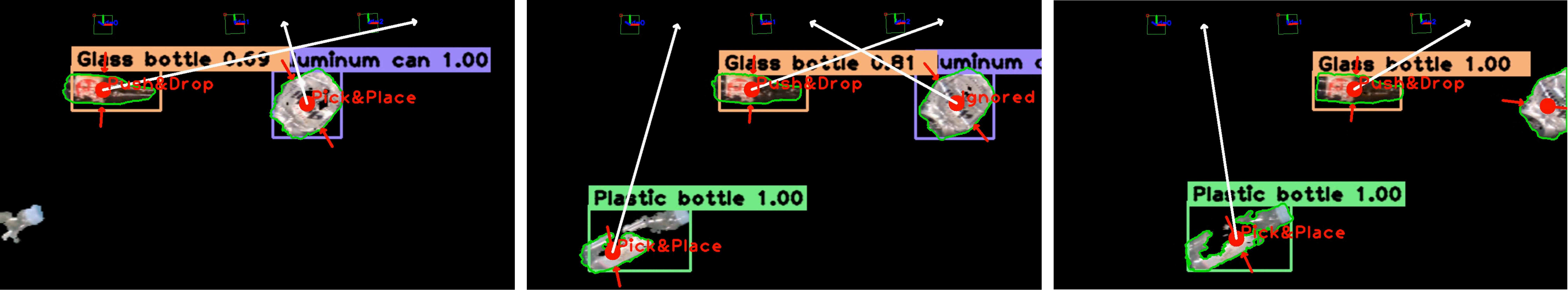}}
  \caption{\small (a) image captured in the waste-sorting scene and (b) the image drawn from the detection results of the waste detector.}
  \figlab{robot-exp}
\end{figure*}
\begin{figure*}[tb]
  \centering
  \includegraphics[width=\linewidth]{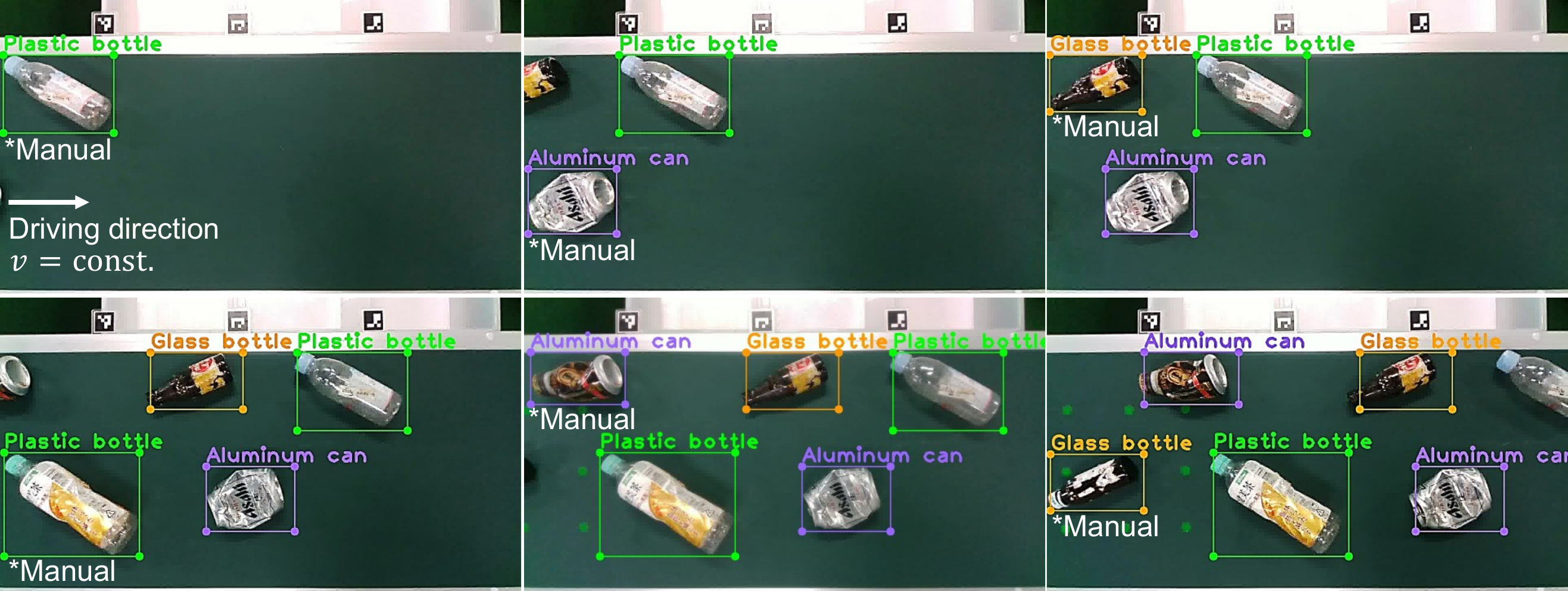}
  \caption{\small Real-world image sequences annotated by humans. *Manual indicates manually annotated bounding boxes. We conducted manual annotation to the video frame in which a new object first appeared. The other images were automatically annotated based on the constant speed of the conveyor and the camera framerate.}
  \figlab{real-collect}
\end{figure*}
\begin{figure*}[tb]
  \centering
  \subcaptionbox{\small{Pick-and-release}}{\includegraphics[width=\textwidth]{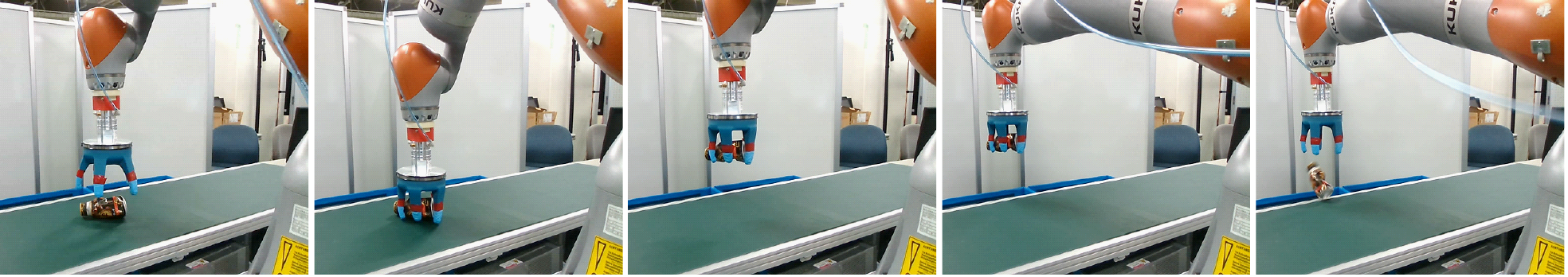}\vspace{-1mm}}
  \subcaptionbox{\small{Push-and-drop}}{\includegraphics[width=\textwidth]{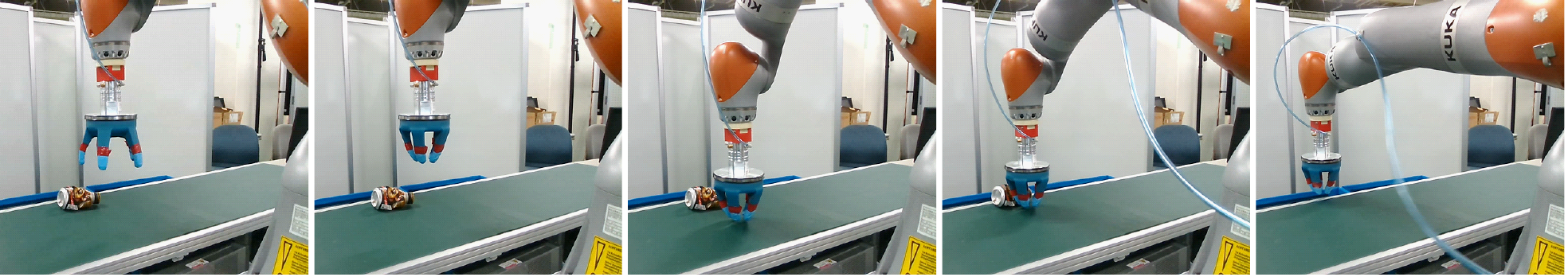}\vspace{-1mm}}
  \caption{\small Two types of manipulation implemented to a waste-sorting robot.}
  \figlab{robot-exp-snaps}
\end{figure*}
\begin{table}[tb]
    \caption{\small Necessary images for adaptation methods.}
    \centering
    \begin{tabular}{ll} \toprule
        \hfil{Method}\hfil & \hfil{Necessary images}\hfil \\ \midrule
        {Image scaling} & {One image pair including a calibration board} \rule[0pt]{0pt}{8pt} \\
        {Background synthesis} & {One background image} \rule[0pt]{0pt}{8pt} \\
        {Color correction} & {One image pair including a color checker} \rule[0pt]{0pt}{8pt} \\ 
        {Seamless cloning} & {One background image} \rule[0pt]{0pt}{8pt} \\ 
        {Histogram matching} & {One object image captured in a real scene} \rule[0pt]{0pt}{8pt} \\ \bottomrule
    \end{tabular}
    \tablab{required-params}
\end{table}
\begin{table}[tb]
    \caption{\small Calculated values of EMD between the reference image (captured in the real scene) and processed images in the training sets. The histogram comparison was conducted in the RGB color space. The values that indicate the highest similarity are shown in bold.}
    \centering
    \begin{tabular}{lp{17mm}p{17mm}p{17mm}} \toprule
        & \multicolumn{3}{c}{{Object category}} \\ \cmidrule(r){2-4}
        {Training set} & \multicolumn{1}{c}{{Aluminum can}} & \multicolumn{1}{c}{{Glass bottle}} & \multicolumn{1}{c}{{Plastic bottle}} \\ \midrule
        {Original} & \hfil{5.86e-1}\hfil & \hfil{8.45e-1}\hfil & \hfil{1.59e0}\hfil \\
        {BS} & \hfil{7.55e-1}\hfil & \hfil{9.41e-1}\hfil & \hfil{1.88e0}\hfil \\
        {BS+CC} & \hfil{8.65e-1}\hfil & \hfil{6.50e-1}\hfil & \hfil{1.77e0}\hfil \\ 
        {SC} & \hfil{2.62e0}\hfil & \hfil{2.10e0}\hfil & \hfil{4.82e0}\hfil \\ 
        {BS+HM} & \hfil{7.36e-3}\hfil & \hfil{5.04e-3}\hfil & \hfil{5.25e-3}\hfil \\ 
        {BS+HM+EQ$^\dagger$} & \hfil{\textbf{6.27e-3}}\hfil & \hfil{\textbf{4.67e-3}}\hfil & \hfil{\textbf{3.96e-3}}\hfil \\\bottomrule
    \end{tabular}
    \begin{tablenotes}
    \item[a]\footnotesize{$^\dagger$ Proposed method in this study.}
    \end{tablenotes}
    \tablab{emd}
\end{table}
\begin{table}[tb]
    \caption{\small Calculated values of BD between the reference image (captured in the real scene) and processed images in the training sets. The histogram comparison was conducted in the RGB color space. The values that indicate the highest similarity are shown in bold.}
    \centering
    \begin{tabular}{lp{17mm}p{17mm}p{17mm}} \toprule
        & \multicolumn{3}{c}{{Object category}} \\ \cmidrule(r){2-4}
        {Training set} & \multicolumn{1}{c}{{Aluminum can}} & \multicolumn{1}{c}{{Glass bottle}} & \multicolumn{1}{c}{{Plastic bottle}} \\ \midrule
        {Original} & \hfil{0.381}\hfil & \hfil{0.425}\hfil & \hfil{0.400}\hfil \\
        {BS} & \hfil{0.436}\hfil & \hfil{0.476}\hfil & \hfil{0.445}\hfil \\
        {BS+CC} & \hfil{0.403}\hfil & \hfil{0.419}\hfil & \hfil{0.428}\hfil \\ 
        {SC} & \hfil{0.454}\hfil & \hfil{0.493}\hfil & \hfil{0.493}\hfil \\ 
        {BS+HM} & \hfil{0.430}\hfil & \hfil{0.445}\hfil & \hfil{0.467}\hfil \\ 
        {BS+HM+EQ$^\dagger$} & \hfil{\textbf{0.220}}\hfil & \hfil{\textbf{0.245}}\hfil & \hfil{\textbf{0.193}}\hfil \\\bottomrule
    \end{tabular}
    \begin{tablenotes}
    \item[a]\footnotesize{$^\dagger$ Proposed method in this study.}
    \end{tablenotes}
    \tablab{bd}
\end{table}

\figref{color-match-result} shows histograms in the RGB color space of the images in \figref{color-match}. The histogram distributions in the RGB color space of the target image (Target) and the converted image (Matched) are visually similar after applying HM.

To conduct a quantitative evaluation, the distance between two histogram distributions were evaluated using \textit{earth--mover's distance} (EMD)~\cite{EMD2000} and \textit{Bhattacharyya distance} (BD)~\cite{Bhattacharyya1943}.

To evaluate the image similarity with the object image captured in the real scene, we calculated the histogram distributions of the four types, which include the original, BS, BS+CC, SC, and BS+HM. 

The effects of the proposed method, BS+HM+EQ, were compared to those of BS+HM, HM, and BS, which are derivatives of the proposed method.
We also compared the comparative methods BS+CC and SC as other color adjustment methods. 

The calculated values of the EMD and BD in the RGB color space are shown in \tabref{emd} and \tabref{bd}.
To compare the images to the object images captured in the real scene, we used those cropped by the bounding boxes as shown in~\figref{both-proc} in red boxes.

The result of the CC shows that the EMD and BD are larger compared with the result of HM.
In the case of the CC, the homography transformation matrix in the RGB color space must be calculated using source and target images, including the color checker shown in~\figref{checker}. 
On the other hand, because the source shown in~\figref{color-match} is converted to become similar to the target shown in~\figref{color-match}, for HM, a higher similarity was achieved.

The calculated values of the EMD and BD suggests that the similarity of the image was largely improved by applying HM, including the area translucent to the back of the object or the plastic bottle's cap. 
This is because the appearance as improved to approximate the target image.
It also suggests that the BS+HM+EQ provided the highest similarity.

\subsection{Detection Accuracy} \label{detection_result}
\begin{table}[tb]
    \caption{\small F-scores of waste category detection using DL-based waste detector trained using each training set [\%]. Mean indicates the mean values of F-score in the three object categories.}
    \centering
    \begin{tabular}{p{23mm}p{10mm}p{10mm}p{10mm}p{10mm}} \toprule
        & \multicolumn{3}{c}{{Object category}} \\ \cmidrule(r){2-4}
        \hfil{Training set}\hfil & \hfil{AC$^{\rm *a}$}\hfil & \hfil{GB$^{\rm *b}$}\hfil & \hfil{PB$^{\rm *c}$}\hfil & \hfil{Mean}\hfil \rule[0pt]{0pt}{8pt} \\ \midrule
        {1. Original} & \hfil{43}\hfil & \hfil{76}\hfil & \hfil{2.0}\hfil & \hfil{40}\hfil \\
        {2. BS} & \hfil{57}\hfil & \hfil{45}\hfil & \hfil{34}\hfil & \hfil{45}\hfil \\
        {3. BS+CC} & \hfil{19}\hfil & \hfil{51}\hfil & \hfil{23}\hfil & \hfil{31}\hfil \\
        {4. SC} & \hfil{14}\hfil & \hfil{51}\hfil & \hfil{10}\hfil & \hfil{25}\hfil \\
        {5. BS+HM} & \hfil{17}\hfil & \hfil{59}\hfil & \hfil{13}\hfil & \hfil{30}\hfil \\
        {6. BS+HM+EQ} & \hfil{22}\hfil & \hfil{64}\hfil & \hfil{28}\hfil & \hfil{38}\hfil \\
        {7. Mixed (1,2,6)} & \hfil{54}\hfil & \hfil{53}\hfil & \hfil{31}\hfil & \hfil{46}\hfil \\
        {8. Real with 7$^\dagger$} & \hfil{72}\hfil & \hfil{89}\hfil & \hfil{75}\hfil & \hfil{79}\hfil \\ \bottomrule
    \end{tabular}
    \begin{tablenotes}
      \item[a]\footnotesize{$^{\rm *a}$ AC is the abbreviation of aluminum can.}
      \item[b]\footnotesize{$^{\rm *b}$ GB is the abbreviation of glass bottle.}
      \item[c]\footnotesize{$^{\rm *c}$ PB is the abbreviation of plastic bottle.}
    \end{tablenotes}
    \begin{tablenotes}
    \item[a]\footnotesize{$^\dagger$ Proposed method in this study.}
    \end{tablenotes}
    \tablab{detection-accuracy}
\end{table}

\tabref{detection-accuracy} shows mean values of detection accuracy for the three target-object categories. As an accuracy metric, we calculated the mean F-score when the IoU threshold was set to 0.5.
We also calculated the F-score using detection results with a confidence value higher than 0.5.
Using a training dataset automatically generated by the proposed method, detection was performed using a waste detector with a trained model of the \textit{single shot multibox detector} (SSD)~\cite{SSD}.
SSD is a general object detector with a convolutional neural--network architecture that learns different anchor boxes. \figref{robot-exp} shows the detection results.

The original shows the result of using 59,400 (120 object-orientation patterns $\times$ 5 viewpoint patterns $\times$ 99 objects) images captured with a green screen shown in \figref{collection-system}. 
BS, BS+CC, SC, BS+HM, and BS+HM+EQ show image training sets subjected to BS only, BS and CC, SC, BS and HM; and BS+HM with EQ, respectively. 
Mixed show the training set that we randomly collected images from the three sets of Original, BS, and BS+HM+EQ.
All the training sets include 59,400 images.

The last set (Real with 7) is a mixed training set that includes the Mixed and 80 images recorded in the real scene, as shown in~\figref{real-collect}. 
The conveyor moves at a constant speed in one direction. Thus, if the image acquisition frequencies of the camera are aligned, the object positions in the images can be shifted at a constant interval. 
Therefore, if we apply manual annotation to only the images of the first frames appearing in the video, we can obtain the image sequence annotated by moving the bounding boxes.
We collected the 80 images from two videos in the waste-sorting scene in this manner.
To improve the quickness of video annotation, in a future work, we plan to use automatic video annotation methods~\cite{Vondrick2011,Kavasidis2014}.

The detection results shown in~\tabref{detection-accuracy} suggest that Mixed provided the highest accuracy of training without images recorded in the waste-sorting environment in the training sets except Real with 7. 
Therefore, our experimental results demonstrate that the accuracy of the waste detector can be improved by applying the aforementioned object scaling, HM with EQ and BS to reduce the differences from the waste-sorting environment.
Surprisingly, the detector with the BS-only dataset showed the almost same accuracy as did Mixed. 
The comparison for these detection accuracies should be done in the future using the backgrounds of various waste-sorting environments.

By adding the small real-world image dataset including the 80 images, we achieved the highest accuracies of detection, even when the number of items in the dataset was small.
The small real-world image dataset not only significantly outperformed the other in terms of accuracy, but the images were also quickly collected.
The time needed to capture a video was about 1 min, and the time needed to annotate only six objects in the six images was about 2 min. This was about 3 min total.

\subsection{Feasibility of Robotic Waste Sorter} \label{feasibility}
\figref{robot-exp-snaps} shows the process by the sorting robot.
In this study, the virtual CoM was calculated as the centroid of the object silhouette extracted from the depth image when the object was viewed from directly above (red dots shown in \figref{robot-exp}(b)). 
The grasp positions during pick-and-release were determined as a straight line on the object silhouette passing through the center of mass perpendicular to the principal axis, as drawn by the red arrows in \figref{robot-exp}(b). 

While sorting manipulation of the waste items by a robot, we evaluated whether the robot succeeded in sorting the waste detected on the conveyor. The success rates of 10 trials of each sorting manipulation for each object category are shown in \tabref{success-rate}.
The results indicate that the pick-and-release operation provided a highly accurate sorting manipulation compared with push-and-drop. 
The average time taken in the 10 trials to finish the push-and-drop operation was 3.3~s, although the time in the case of the pick-and-release operation took 5.2~s.
Our algorithm reduced the time required for manipulation by simplifying the manipulation process.

As examples of failures in the pick-and-release, we confirmed cases where a large object did not fit in the grasping area, cases where the grasping failed due to an error of the estimated virtual CoM, and cases where the released object by placing motion did reach the target bin.

First, we must consider another grasping method based on the gripper's grasping area and target object size.
In the case of container and packaging waste, there are many large slender objects. 
Thus, we need another grasping method in which the thinnest part can be sandwiched between two of the five fingers.
Second, we require object segmentation~\cite{Felzenszwalb2004,Xiong2019} or foreground extraction~\cite{Rother2004,Scholkopf2006,Kim2007} methods that use color information, because the silhouette sometimes cannot be generated, owing object--region extraction errors by the depth image.
Third, the target garbage item was not put into the target bin, because the acceleration of the robot arm sent it flying over top. 
We should not slow the robot arm motion even for this case, owing to the low agility of manipulation. 
We instead require a particular a release motion by a robot arm that accounts for acceleration. 
The pick-and-place for dynamic objects~\cite{Cowley2013} could also achieve highly accurate sorting.

As an example of failures in the push-and-drop, we first confirmed cases where the gripper's fingers could not make good contact with the sides of the target objects.
To ensure reliable contact for pushing, we must consider the waste-item shape and the orientation of the gripper.

Second, we confirmed a case in which the target object overshot the bin and another where the target object was too heavy to exit the conveyor.
There was also a case in which the target object only rotated after pushing.
Therefore, we need to generate a pushing motion based on the target object weight and shape~\cite{Zhou2016}.
\begin{table}[tb]
    \centering
    \caption{\small Results of the sorting manipulation. Each element shows success rate [\%] in each 10 trials.}
    \begin{tabular}{p{23mm}p{10mm}p{10mm}p{10mm}p{10mm}} \toprule
        & \multicolumn{3}{c}{{Object category}} \\ \cmidrule(r){2-4}
        \multicolumn{1}{c}{Method} & \multicolumn{1}{c}{{AC$^{\rm *a}$}} & \multicolumn{1}{c}{{GB$^{\rm *b}$}} & \multicolumn{1}{c}{{PB$^{\rm *c}$}} & \multicolumn{1}{c}{{Mean}} \\ \midrule
        {Pick-and-release} & \hfil{80}\hfil & \hfil{70}\hfil & \hfil{60}\hfil & \hfil{70}\hfil \rule[0pt]{0pt}{10pt} \\
        {Push-and-drop} & \hfil{60} & \hfil{50}\hfil & \hfil{50}\hfil & \hfil{57}\hfil \rule[0pt]{0pt}{10pt} \\ \bottomrule
    \end{tabular}
        \begin{tablenotes}
          \item[a]\footnotesize{$^{\rm *a}$ AC is the abbreviation of aluminum can.}
          \item[b]\footnotesize{$^{\rm *b}$ GB is the abbreviation of glass bottle.}
          \item[c]\footnotesize{$^{\rm *c}$ PB is the abbreviation of plastic bottle.}
        \end{tablenotes}
    \tablab{success-rate}
\end{table}
\section{Discussion on Future Issues}
\subsection{Ensuring High Consistencies of Illumination and Geometry}
The purpose of this study, apart from reducing the time required for dataset collection, was to achieve a highly accurate detector. 
Within this context, for the consistency of illumination, we proposed a method that matches only the luminance distribution information of the image without considering a camera-response function~\cite{Takamatsu2008} and the distribution of the light source~\cite{Sato2003,Hara2005} in the different environments.
In reality, these optical models must be considered when obtaining more realistic images that are similar to real-world ones. 
However, estimation methods requiring less labor are needed.

In terms of geometric consistency, in this study, only the distance from the camera to the object was considered.
However, a 3D model is needed to transform the geometry more precisely.
One idea for generating realistic images via a 3D model requires free viewpoint image synthesis based on 3D shape reconstruction methods, such as \textit{Space carving}~\cite{SpaceCarving}, and a geometric registration and an alignment using an RGBD video~\cite{RobustReconst}.

\subsection{Precise Annotation}
\figref{problem-annotate} shows the four cases that had difficulty annotating collected images, especially for cases of difficult object-region extraction.
The problematic images shown in~\figref{problem-annotate} include an object adhered to foreign substances, a semi-transparent object, a shadow under the object, and a green object.

The foreign substances shown in~\figref{problem-annotate}(a) needs to be removed from the target object, because the waste detector is not designed to recognize this part. 
Consequently, the waste-sorting robot cannot grasp and push the part.
\figref{problem-annotate}(b) shows a misannotated semi-transparent object.
For the automatic annotation, we could in the future use another method that does not rely exclusively on optical information.
As shown in~\figref{problem-annotate}(c), because it may be difficult to distinguish a boundary from a shadow, object region extraction may fail.
In a future work, it will be necessary to improve the algorithm so that it is robust to shading by referring to illumination estimation methods~\cite{Finlayson2006,Panagopoulos2011} and DL-based shadow detection and removal methods~\cite{Nguyen2017,Qu2017}.
To avoid difficulty of region extraction caused by similar colors, as shown in~\figref{problem-annotate}(d), background coloring should be considered.

\begin{figure}[tb]
  \centering
  \subcaptionbox{\small{Foreign substances adhered} }{\includegraphics[width=\linewidth]{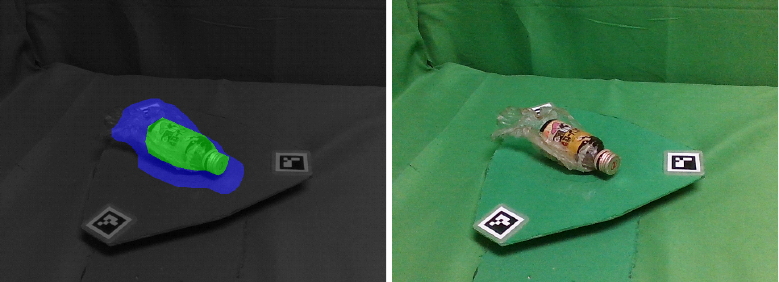}\vspace{-1mm}}
  \subcaptionbox{\small{Semi-transparent object}}{\includegraphics[width=\linewidth]{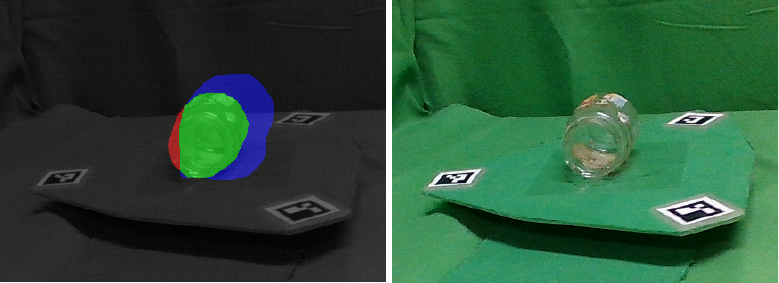}\vspace{-1mm}}
  \subcaptionbox{\small{Shadow misannotated}}{\includegraphics[width=\linewidth]{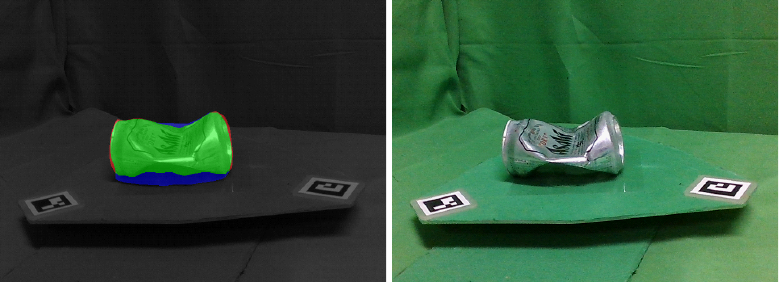}\vspace{-1mm}}
  \subcaptionbox{\small{Green object} }{\includegraphics[width=\linewidth]{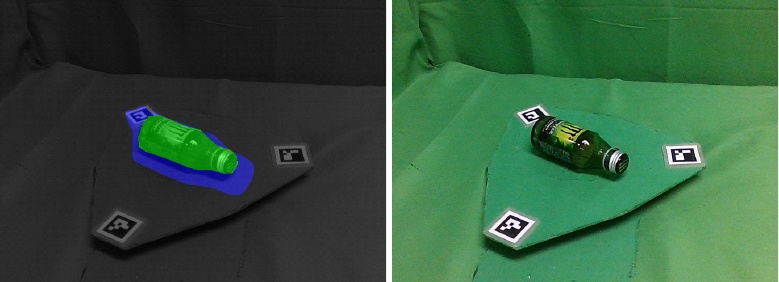}\vspace{-1mm}}
  \caption{\small Problematic images difficult to annotate. The coloring in each left image is the same as that of~\figref{anno-result}.}
  \figlab{problem-annotate}
\end{figure}

\section{Conclusion}
In this study, to achieve an agile waste-sorting method, we first proposed two types of manipulation and a selection algorithm based on time constraints of the conveyed waste.

Second, to reduce the time required for capturing object images and annotations, we developed a robotic training dataset collection system using a small hand--eye robot and a rotating stage.

Third, to fill the gap between the generated image set and the one captured from a waste-sorting scene, we provided an image adaptation method. 

In our experiment, we successfully automatically generated a training set using the proposed robotic training dataset collection system.
To train the waste detector, we applied the proposed adaptation method, including histogram matching with histogram equalization, background synthesis, and object scaling of the collected dataset. 
Finally, the waste detector performed waste detection, and the robotic waste-sorting system successfully performed pick-and-release and push-and-drop in a real work environment. 

The dataset collection time was reduced to at least 1\% or less of the previously proposed automatic dataset collection method.
We verified that the waste detector could detect target waste items (\ie~aluminum cans, glass bottles, and plastic bottles) in a waste-sorting environment. 
As a result, the mean F-score for all objects was about 46\%, and the accuracy was higher than the method lacking adaptation methods.
We achieved a highly accurate detector trained with the training set, including the proposed dataset and a small dataset captured in a real scene. The mean value of the F-score in the three object categories was about 79\%.

The robot successfully demonstrated the two types of manipulation at a success rate greater than 61\%. 
The push-and-drop of the graspless manipulation more quickly performed the sorting manipulation for one object than did the pick-and-release method by 1.9~s.
The average time taken in the 10 trials to finish the push-and-drop operation was 3.3~s, although the time in the case of the pick-and-release operation took 5.2~s.

As our future works, we consider other system configurations: the one system using multiple cameras to more accurately detect the waste items and the one system using other flexible endeffectors like brush-shaped gripper to more robustly manipulate the irregular-shaped waste items.

\section*{Acknowledgement}
This work was supported by the projects, JPNP14004 and JPNP20012 subsidized by the New Energy and Industrial Technology Development Organization (NEDO).

\bibliographystyle{IEEEtran}
\bibliography{reference.bib}

\begin{thebibliography}{10}
\providecommand{\url}[1]{#1}
\csname url@samestyle\endcsname
\providecommand{\newblock}{\relax}
\providecommand{\bibinfo}[2]{#2}
\providecommand{\BIBentrySTDinterwordspacing}{\spaceskip=0pt\relax}
\providecommand{\BIBentryALTinterwordstretchfactor}{4}
\providecommand{\BIBentryALTinterwordspacing}{\spaceskip=\fontdimen2\font plus
\BIBentryALTinterwordstretchfactor\fontdimen3\font minus
  \fontdimen4\font\relax}
\providecommand{\BIBforeignlanguage}[2]{{%
\expandafter\ifx\csname l@#1\endcsname\relax
\typeout{** WARNING: IEEEtran.bst: No hyphenation pattern has been}%
\typeout{** loaded for the language `#1'. Using the pattern for}%
\typeout{** the default language instead.}%
\else
\language=\csname l@#1\endcsname
\fi
#2}}
\providecommand{\BIBdecl}{\relax}
\BIBdecl

\bibitem{Gundupalli2017}
S.~P. Gundupalli, S.~Hait, and A.~Thakur, ``A review on automated sorting of
  source-separated municipal solid waste for recycling,'' \emph{Waste
  Management}, vol.~60, pp. 56--74, 2017.

\bibitem{Chahine2018}
K.~Chahine and B.~Ghazal, ``Automatic sorting of solid wastes using sensor
  fusion,'' \emph{International Journal of Engineering and Technology}, vol.~9,
  no.~6, pp. 4408--4414, 2018.

\bibitem{Industry4Colombia}
N.~Barrero, D.~Galvis, and C.~Martinez, ``Industrial robots for waste
  separation tasks: An approach to industry 4.0 in colombia,'' \emph{{\rm in}
  Proc. 9th International Conference on Production Research-Americas}, 2018.

\bibitem{ReviewWasteRobot}
R.~Sarc, A.~Curtis, L.~Kandlbauer, K.~Khodier, K.~E. Lorber, and R.~Pomberger,
  ``Digitalisation and intelligent robotics in value chain of circular economy
  oriented waste management – a review,'' \emph{Waste Management}, vol.~95,
  pp. 476--492, 2019.

\bibitem{RecyclingRobots}
T.~Gibson, ``{Recycling Robots},'' \emph{Mechanical Engineering}, vol. 142,
  no.~01, pp. 32--37, 2020.

\bibitem{Lukka2014}
T.~J. Lukka, T.~Tossavainen, J.~V. Kujala, and R.~Tapani, ``{ZenRobotics
  Recycler} – robotic sorting using machine learning,'' \emph{{\rm in} SBS},
  pp. 1--8, 2014.

\bibitem{FastRCNN}
R.~Girshick, ``Fast {R-CNN},'' \emph{{\rm in} ICCV}, pp. 1440–--1448, 2015.

\bibitem{FasterRCNN}
S.~Ren, K.~He, R.~Girshick, and J.~Sun, ``Faster {R-CNN}: Towards real-time
  object detection with region proposal networks,'' \emph{{\rm in} NIPS}, pp.
  91–--99, 2015.

\bibitem{SSD}
W.~Liu, D.~Anguelov, D.~Erhan, C.~Szegedy, S.~Reed, C.-Y. Fu, and A.~C. Berg,
  ``{SSD}: Single shot multibox detector,'' \emph{{\rm in} ECCV}, pp. 21--37,
  2016.

\bibitem{YOLOv3}
J.~Redmon and A.~Farhadi, ``{Yolov3}: An incremental improvement,'' \emph{CoRR,
  abs/1804.02767}, 2018.

\bibitem{M2Det}
Q.~Zhao, T.~Sheng, Y.~Wang, Z.~Tang, Y.~Chen, L.~Cai, and H.~Ling, ``{M2Det}: A
  single-shot object detector based on multi-level feature pyramid network,''
  \emph{{\rm in} AAAI}, pp. 9259--9266, 2019.

\bibitem{FCOS}
T.~Zhi, C.~Shen, H.~Chen, and T.~He, ``{FCOS}: Fully convolutional one-stage
  object detection,'' \emph{{\rm in} ICCV}, pp. 9626--9635, 2019.

\bibitem{EfficientDet}
M.~Tan, R.~Pang, and Q.~V. Le, ``{EfficientDet}: Scalable and efficient object
  detection,'' \emph{{\rm in} CVPR}, pp. 10\,781--10\,790, 2020.

\bibitem{ImageNet}
J.~{Deng}, W.~{Dong}, R.~{Socher}, L.~{Li}, {Kai Li}, and {Li Fei-Fei},
  ``{ImageNet}: A large-scale hierarchical image database,'' \emph{{\rm in}
  CVPR}, pp. 248--255, 2009.

\bibitem{PascalVOC}
M.~Everingham, S.~M.~A. Eslami, L.~Van~Gool, C.~K.~I. Williams, and J.~Winn,
  ``The pascal visual object classes challenge: A retrospective,''
  \emph{International Journal of Computer Vision}, vol. 111, no.~1, pp.
  98--136, 2015.

\bibitem{MSCOCO}
T.-Y. Lin, M.~Maire, B.~Serge, H.~James, P.~Perona, D.~Ramanan, P.~Dollár, and
  C.~L. Zitnick, ``Microsoft {COCO}: Common objects in context,'' \emph{{\rm
  in} ECCV}, pp. 740--755, 2014.

\bibitem{CityScapes}
M.~Cordts, M.~Omran, S.~Ramos, T.~Rehfeld, M.~Enzweiler, R.~Benenson,
  U.~Franke, S.~Roth, and B.~Schiele, ``The cityscapes dataset for semantic
  urban scene understanding,'' \emph{{\rm in} CVPR}, pp. 3213--3223, 2016.

\bibitem{ReviewDL}
Z.-Q. Zhao, P.~Zheng, S.-T. Xu, and X.~Wu, ``Object detection with deep
  learning: A review,'' \emph{IEEE Transactions on Neural Networks and Learning
  Systems}, vol.~30, no.~11, pp. 3212--3232, 2019.

\bibitem{RA-L2019}
T.~Kiyokawa, K.~Tomochika, J.~Takamatsu, and T.~Ogasawara, ``Fully automated
  annotation with noise-masked visual markers for deep-learning-based object
  detection,'' \emph{Robotics and Automation Letters}, vol.~4, no.~2, pp.
  1972--1977, 2019.

\bibitem{Kato2001}
H.~Kato and M.~Billinghurst, ``Marker tracking and hmd calibration for a
  video-based augmented reality conferencing system,'' \emph{{\rm in} IWAR},
  pp. 85--94, 1999.

\bibitem{AdvancedRobotics2019}
T.~Kiyokawa, K.~Tomochika, J.~Takamatsu, and T.~Ogasawara, ``Efficient
  collection and automatic annotation of real-world object images by taking
  advantage of post-diminished multiple visual markers,'' \emph{Advanced
  Robotics}, vol.~33, no.~24, pp. 1264--1280, 2019.

\bibitem{ZenRobot2015}
J.~V. Kujala, T.~J. Lukka, and H.~Holopainen, ``Classifying and sorting
  cluttered piles of unknown objects with robots: A learning approach,''
  \emph{{\rm in} IROS}, pp. 971--978, 2016.

\bibitem{GrasplessAiyama}
Y.~Aiyama, M.~Inaba, and H.~Inoue, ``Pivoting: A new method of graspless
  manipulation of object by robot fingers,'' \emph{{\rm in} IROS}, pp.
  136--143, 1993.

\bibitem{GrasplessMaeda}
Y.~Maeda and T.~Arai, ``Planning of graspless manipulation by a multifingered
  robot hand,'' \emph{Advanced Robotics}, vol.~19, no.~5, pp. 501–--521,
  2005.

\bibitem{Nikhil2015}
N.~Chavan-Dafle and A.~Rodriguez, ``Prehensile pushing: In-hand manipulation
  with push-primitives,'' in \emph{IROS}, 2015, pp. 6215--6222.

\bibitem{NonprehensileManip}
M.~T. Mason, ``{Progress in Nonprehensile Manipulation},'' \emph{The
  International Journal of Robotics Research}, vol.~18, no.~11, pp. 1129--1141,
  1999.

\bibitem{DynamicNonPrehensile}
K.~M. Lynch and M.~T. Mason, ``Dynamic nonprehensile manipulation:
  Controllability, planning, and experiments,'' \emph{The International Journal
  of Robotics Research}, vol.~18, no.~1, pp. 64--92, 1999.

\bibitem{Plastics}
S.~R. Ahmad, ``A new technology for automatic identification and sorting of
  plastics for recycling,'' \emph{Environmental Technology}, vol.~25, pp.
  1143--1149, 2004.

\bibitem{OpticalSensor}
J.~Huang, T.~Pretz, and Z.~Bian, ``Intelligent solid waste processing using
  optical sensor based sorting technology,'' \emph{{\rm in} CISP}, pp.
  1657--1661, 2010.

\bibitem{LaserSensor}
H.~Jull, J.~Bier, R.~K\"{u}nnemeyer, and P.~Schaare, ``Classification of
  recyclables using laser-induced breakdown spectroscopy for waste
  management,'' \emph{Spectroscopy Letters}, vol.~51, no.~6, pp. 257--265,
  2018.

\bibitem{Thermalewaste}
S.~P. Gundupalli, S.~Hait, A.~Thakur, and A.~Trivedi, ``Classification of
  recyclables from e-waste stream using thermal imaging-based technique,''
  \emph{Urbanization Challenges in Emerging Economies: Energy and Water
  Infrastructure}, pp. 67--78, 2018.

\bibitem{ThermalMetal}
S.~P. Gundupalli, S.~Hait, and A.~Thakur, ``Classification of metallic and
  non-metallic fractions of e-waste using thermal imaging-based technique,''
  \emph{Process Safety and Environmental Protection}, vol. 118, pp. 32--39,
  2018.

\bibitem{Mao2021}
W.-L. Mao, W.-C. Chen, C.-T. Wang, and Y.-H. Lin, ``Recycling waste
  classification using optimized convolutional neural network,''
  \emph{Resources, Conservation and Recycling}, vol. 164, p. 105132, 2021.

\bibitem{BinYan2017}
L.~Binyan, W.~Yan-bo, C.~Zhihong, L.~Jia-yu, and L.~Junqin, ``Object detection
  and robotic sorting system in complex industrial environment,'' \emph{{\rm
  in} CAC}, pp. 7277--7281, 2017.

\bibitem{ButtonCellBatteries}
H.~Karbasi, A.~Sanderson, A.~Sharifi, and C.~Pop, ``Robotic sorting of used
  button cell batteries: Utilizing deep learning,'' \emph{{\rm in} SusTech},
  pp. 1--6, 2018.

\bibitem{Zhifei2019}
Z.~Zhang, H.~Wang, H.~Song, S.~Zhang, and J.~Zhang, ``Industrial robot sorting
  system for municipal solid waste,'' \emph{Intelligent Robotics and
  Applications}, pp. 342--353, 2019.

\bibitem{DeepLearning}
Y.~LeCun, Y.~Bengio, and G.~Hinton, ``Deep learning,'' \emph{Nature}, vol. 521,
  pp. 436--44, 2015.

\bibitem{OnGlass}
J.~Bai, S.~Lian, Z.~Liu, K.~Wang, and D.~Liu, ``Deep learning based robot for
  automatically picking up garbage on the grass,'' \emph{IEEE Transactions on
  Consumer Electronics}, vol.~64, no.~3, pp. 382--389, 2018.

\bibitem{GarbageDetChina}
C.~Zhihong, Z.~Hebin, W.~Yan, W.~Yanbo, and L.~Binyan, ``Multi-task detection
  system for garbage sorting base on high-order fusion of convolutional feature
  hierarchical representation,'' \emph{{\rm in} 37th Chinese Control
  Conference}, pp. 5426--5430, 2018.

\bibitem{RICAP}
R.~Takahashi, T.~Matsubara, and K.~Uehara, ``{RICAP}: Random image cropping and
  patching data augmentation for deep {CNNs},'' \emph{{\rm in} ACML}, pp.
  786--798, 2018.

\bibitem{RondomErasing}
Z.~Zhong, L.~Zheng, G.~Kang, S.~Li, and Y.~Yang, ``Random erasing data
  augmentation,'' \emph{{\rm in} AAAI}, pp. 13\,001--13\,008, 2020.

\bibitem{AutoAugment}
E.~D. Cubuk, B.~Zoph, D.~Mane, V.~Vasudevan, and Q.~V. Le, ``Autoaugment:
  Learning augmentation strategies from data,'' \emph{{\rm in} CVPR}, pp.
  113--123, 2019.

\bibitem{FastAutoAugment}
S.~Lim, I.~Kim, T.~Kim, C.~Kim, and S.~Kim, ``Fast {AutoAugment},'' \emph{{\rm
  in} NeurIPS}, pp. 6665–--6675, 2019.

\bibitem{ExClick}
D.~P. Papadopoulos, J.~R.~R. Uijlings, F.~Keller, and V.~Ferrari, ``{Extreme
  clicking for efficient object annotation},'' \emph{{\rm in} ICCV}, pp.
  4930--4939, 2017.

\bibitem{DeepExCut}
K.-K. Maninis, S.~Caelles, J.~Pont-Tuset, and L.~Van~Gool, ``Deep extreme cut:
  From extreme points to object segmentation,'' \emph{{\rm in} CVPR}, pp.
  616--625, 2018.

\bibitem{NVIDIAannotation}
H.~Ling, J.~Gao, A.~Kar, W.~Chen, and S.~Fidler, ``Fast interactive object
  annotation with {Curve-GCN},'' \emph{{\rm in} CVPR}, pp. 5257--5266, 2019.

\bibitem{GoogleAnnotation}
R.~Benenson, S.~Popov, and V.~Ferrari, ``Large-scale interactive object
  segmentation with human annotators,'' \emph{{\rm in} CVPR}, pp.
  11\,700--11\,709, 2019.

\bibitem{EasyLabel}
M.~Suchi, T.~Patten, D.~Fischinger, and M.~Vincze, ``{EasyLabel}: A
  semi-automatic pixel-wise object annotation tool for creating robotic {RGB-D}
  datasets,'' \emph{{\rm in} ICRA}, pp. 6678--6684, 2019.

\bibitem{SemiautoLabel}
D.~De~Gregorio, A.~Tonioni, G.~Palli, and L.~Di~Stefano, ``Semiautomatic
  labeling for deep learning in robotics,'' \emph{IEEE Trans. on Automation
  Science and Engineering}, vol.~17, no.~2, pp. 611--620, 2020.

\bibitem{SemiautoTrainGen}
S.~Akizuki and M.~Hashimoto, ``Semi-automatic training data generation for
  semantic segmentation using {6DoF} pose estimation,'' \emph{{\rm in} VISAPP},
  pp. 607--613, 2019.

\bibitem{Tzeng2017}
E.~Tzeng, J.~Hoffman, K.~Saenko, and T.~Darrell, ``Adversarial discriminative
  domain adaptation,'' \emph{{\rm in} CVPR}, pp. 7167--7176, 2017.

\bibitem{Raghuraman2011}
R.~Gopalan, R.~Li, and R.~Chellappa, ``Domain adaptation for object
  recognition: An unsupervised approach,'' \emph{{\rm in} ICCV}, pp. 999--1006,
  2011.

\bibitem{SynthesizingRSS17}
G.~Georgakis, A.~Mousavian, A.~C. Berg, and J.~Ko\v{s}eck\'{a}, ``Synthesizing
  training data for object detection in indoor scenes,'' \emph{{\rm in} RSS},
  2017.

\bibitem{Hsu2019}
H.-K. Hsu, W.-C. Hung, H.-Y. Tseng, C.-H. Yao, Y.-H. Tsai, M.~Singh, and M.-H.
  Yang, ``Progressive domain adaptation for object detection,'' \emph{{\rm in}
  CVPR}, pp. 1--5, 2019.

\bibitem{Inoue2018}
N.~Inoue, R.~Furuta, T.~Yamasaki, and K.~Aizawa, ``Cross-domain
  weakly-supervised object detection through progressive domain adaptation,''
  \emph{{\rm in} CVPR}, pp. 5001--5009, 2018.

\bibitem{Luo2019}
Y.~Luo, L.~Zheng, T.~Guan, J.~Yu, and Y.~Yang, ``Taking a closer look at domain
  shift: Category-level adversaries for semantics consistent domain
  adaptation,'' \emph{{\rm in} CVPR}, pp. 2507--2516, 2019.

\bibitem{AlphaMatting}
K.~He, J.~Sun, and X.~Tang, ``Fast matting using large kernel matting laplacian
  matrices,'' \emph{{\rm in} CVPR}, pp. 2165--2172, 2010.

\bibitem{Germer2020}
T.~Germer, T.~Uelwer, S.~Conrad, and S.~Harmeling, ``{PyMatting}: A python
  library for alpha matting,'' \emph{Journal of Open Source Software}, vol.~5,
  no.~54, p. 2481, 2020.

\bibitem{DIP2nd}
R.~C. Gonzalez and R.~E. Woods, \emph{Digital Image Processing 2nd
  Edition}.\hskip 1em plus 0.5em minus 0.4em\relax Addison-Wesley, 2001, ch.~3,
  pp. 94--102.

\bibitem{AlgApp}
R.~Szeliski, \emph{Computer Vision: Algorithms and Applications}.\hskip 1em
  plus 0.5em minus 0.4em\relax Springer-Verlag London, 2011, ch.~2.

\bibitem{CLAHE1994}
K.~Zuiderveld, ``Contrast limited adaptive histograph equalization,''
  \emph{Graphic Gems IV. San Diego: Academic Press Professional}, pp.
  474–--485, 1994.

\bibitem{ChittaRAM2012}
S.~Chitta, I.~Sucan, and S.~Cousins, ``Moveit!'' \emph{IEEE Robotics Automation
  Magazine}, vol.~19, no.~1, pp. 18--19, 2012.

\bibitem{ArucoMain}
S.~Garrido-Jurado, R.~Mu\={n}oz-Salinas, F.~J. Madrid-Cuevas, and
  R.~Medina-Carnicer, ``Generation of fiducial marker dictionaries using mixed
  integer linear programming,'' \emph{Pattern Recognition}, vol.~51, pp.
  481--491, 2016.

\bibitem{ArucoAdditional}
F.~J. Romero-Ramirez, R.~Mu\={n}oz-Salinas, and R.~Medina-Carnicer, ``Speeded
  up detection of squared fiducial markers,'' \emph{Image and Vision
  Computing}, vol.~76, pp. 38--47, 2018.

\bibitem{Wang2020}
Z.~Wang, E.~Wang, and Y.~Zhu, ``Image segmentation evaluation: a survey of
  methods,'' \emph{Artificial Intelligence Review}, vol.~53, pp. 5637–--5674,
  2020.

\bibitem{PlantCV}
J.~C. Berry, N.~Fahlgren, A.~P. Pokorny, R.~Bart, and K.~M. Veley, ``An
  automated, high-throughput method for standardizing image color profiles to
  improve image-based plant phenotyping,'' \emph{PeerJ}, vol.~6, p. e5727,
  2018.

\bibitem{ColorHomography}
H.~Gong, G.~D. Finlayson, and R.~B. Fisher, ``Recoding color transfer as a
  color homography,'' \emph{{\rm in} BMVC}, pp. 17.1--17.11, 2016.

\bibitem{PIE2003}
P.~P\'{e}rez, M.~Gangnet, and A.~Blake, ``Poisson image editing,'' \emph{ACM
  Transactions on Graphics}, vol.~22, no.~3, pp. 313--–318, 2003.

\bibitem{Kakuta2007}
T.~Kakuta, T.~Oishi, and K.~Ikeuchi, ``Real-time soft shadows in mixed reality
  using shadowing planes,'' \emph{{\rm in} MVA}, pp. 195--198, 2007.

\bibitem{Mukaigawa2001}
Y.~Mukaigawa, H.~Miyaki, S.~Mihashi, and T.~Shakunaga, ``Photometric
  image-based rendering for image generation in arbitrary illumination,''
  \emph{{\rm in} ICCV}, pp. 652--659, 2001.

\bibitem{Sato2005}
I.~Sato, M.~Hayashida, F.~Kai, Y.~Sato, and K.~Ikeuchi, ``Fast image synthesis
  of virtual objects in a real scene with natural shadings,'' \emph{Systems and
  Computers in Japan}, vol.~36, no.~14, pp. 102--111, 2005.

\bibitem{Yokoya2015}
F.~Okura, M.~Kanbara, and N.~Yokoya, ``Mixed-reality world exploration using
  image-based rendering,'' \emph{ACM Journal on Computing and Cultural
  Heritage}, vol.~8, no.~2, pp. 9:1--9:26, 2015.

\bibitem{EMD2000}
Y.~Rubner, C.~Tomasi, and L.~J. Guibas, ``The earth mover's distance as a
  metric for image retrieval,'' \emph{International Journal of Computer
  Vision}, vol.~40, no.~2, pp. 99--121, 2000.

\bibitem{Bhattacharyya1943}
A.~Bhattacharyya, ``On a measure of divergence between two statistical
  populations defined by probability distributions,'' \emph{Bulletin of the
  Calcutta Mathematical Society}, vol.~35, p. 99–109, 1943.

\bibitem{Vondrick2011}
C.~Vondrick and D.~Ramanan, ``Video annotation and tracking with active
  learning,'' \emph{{\rm in} NIPS}, pp. 28–--36, 2011.

\bibitem{Kavasidis2014}
I.~Kavasidis, S.~Palazzo, D.~R. Salvo, D.~Giordano, and C.~Spampinato, ``An
  innovative web-based collaborative platform for video annotation,''
  \emph{Multimedia Tools and Applications}, vol.~70, pp. 413--432, 2014.

\bibitem{Felzenszwalb2004}
P.~Felzenszwalb and D.~Huttenlocher, ``Efficient graph-based image
  segmentation,'' \emph{International Journal of Computer Vision}, vol.~59,
  no.~2, pp. 167--181, 2004.

\bibitem{Xiong2019}
B.~Xiong, S.~Jain, and K.~Grauman, ``{Pixel Objectness}: Learning to segment
  generic objects automatically in images and videos,'' \emph{IEEE Transactions
  on Pattern Analysis and Machine Intelligence}, vol.~41, no.~11, pp.
  2677--2692, 2019.

\bibitem{Rother2004}
C.~Rother, V.~Kolmogorov, and A.~Blake, ``{"GrabCut"}: Interactive foreground
  extraction using iterated graph cuts,'' in \emph{ACM SIGGRAPH}, 2004, pp.
  309--314.

\bibitem{Scholkopf2006}
B.~{Schölkopf}, J.~{Platt}, and T.~{Hofmann}, ``Dynamic foreground/background
  extraction from images and videos using random patches,'' in \emph{NIPS},
  2006, pp. 929--936.

\bibitem{Kim2007}
H.~Kim, R.~Sakamoto, I.~Kitahara, T.~Toriyama, and K.~Kogure, ``Robust
  foreground extraction technique using gaussian family model and multiple
  thresholds,'' in \emph{ACCV}, 2007, pp. 758--768.

\bibitem{Cowley2013}
A.~{Cowley}, B.~{Cohen}, W.~{Marshall}, C.~J. {Taylor}, and M.~{Likhachev},
  ``Perception and motion planning for pick-and-place of dynamic objects,'' in
  \emph{IROS}, 2013, pp. 816--823.

\bibitem{Zhou2016}
{Jiaji Zhou}, R.~{Paolini}, J.~A. {Bagnell}, and M.~T. {Mason}, ``A convex
  polynomial force-motion model for planar sliding: Identification and
  application,'' in \emph{ICRA}, 2016, pp. 372--377.

\bibitem{Takamatsu2008}
J.~Takamatsu, Y.~Matsushita, and K.~Ikeuchi, ``Estimating camera response
  functions using probabilistic intensity similarity,'' \emph{{\rm in} CVPR},
  pp. 1--8, 2008.

\bibitem{Sato2003}
S.~Imari, S.~Yoichi, and I.~Katsushi, ``Illumination from shadows,'' \emph{IEEE
  Transactions on Pattern Analysis and Machine Intelligence}, vol.~25, no.~3,
  pp. 290--300, 2003.

\bibitem{Hara2005}
K.~Hara, K.~Nishino, and K.~Ikeuchi, ``Multiple light sources and reflectance
  property estimation based on a mixture of spherical distributions,''
  \emph{{\rm in} ICCV}, pp. 1627--1634, 2005.

\bibitem{SpaceCarving}
K.~N. Kutulakos and S.~M. Seitz, ``A theory of shape by space carving,''
  \emph{International Journal of Computer Vision}, vol.~38, no.~3, pp.
  199–--218, 2000.

\bibitem{RobustReconst}
S.~Choi, Q.-Y. Zhou, and V.~Koltun, ``Robust reconstruction of indoor scenes,''
  \emph{{\rm in} CVPR}, pp. 5556--5565, 2015.

\bibitem{Finlayson2006}
G.~D. {Finlayson}, S.~D. {Hordley}, {Cheng Lu}, and M.~S. {Drew}, ``On the
  removal of shadows from images,'' \emph{IEEE Transactions on Pattern Analysis
  and Machine Intelligence}, vol.~28, no.~1, pp. 59--68, 2006.

\bibitem{Panagopoulos2011}
A.~{Panagopoulos}, C.~{Wang}, D.~{Samaras}, and N.~{Paragios}, ``Illumination
  estimation and cast shadow detection through a higher-order graphical
  model,'' \emph{{\rm in} CVPR}, pp. 673--680, 2011.

\bibitem{Nguyen2017}
V.~Nguyen, T.~F.~Y. Vicente, M.~Zhao, M.~Hoai, and D.~Samaras, ``Shadow
  detection with conditional generative adversarial networks,'' \emph{IEEE
  International Conference on Computer Vision}, pp. 4520–--4528, 2017.

\bibitem{Qu2017}
L.~Qu, J.~Tian, S.~He, Y.~Tan, and R.~W.~H. Lau, ``{DeshadowNet}: A
  multi-context embedding deep network for shadow removal,'' \emph{IEEE
  International Conference on Computer Vision}, pp. 2308–--2316, 2017.

\end{thebibliography}

\end{document}